\documentclass[10pt, a4paper]{article}
\usepackage{inconsolata}
\usepackage{amsmath}
\usepackage{amsfonts}
\usepackage{subfigure}
\usepackage{graphicx}
\usepackage{booktabs}
\usepackage{url}
\usepackage{tabularx}
\usepackage{bigstrut}
\usepackage{multirow}
\usepackage{color}
\usepackage{xcolor}
\usepackage{verbatimbox}
\usepackage{enumitem}
\usepackage{xspace}
\usepackage{gensymb}
\usepackage{caption} 
\usepackage{rotating}
\usepackage{makecell}

\usepackage{microtype}
\usepackage[utf8]{inputenc}
\usepackage[T1]{fontenc}

\usepackage{times}
\usepackage{latexsym}

\usepackage[review]{lrec-coling2024} 

\title{Towards Few-shot Entity Recognition in Document Images: \\A Graph Neural Network Approach Robust to Image Manipulation \\ \vspace*{.5\baselineskip} }

\name{Prashant Krishnan, Zilong Wang, Yangkun Wang, Jingbo Shang } 

\address{University of California, San Diego \\ 
pvaidyanathan@ucsd.edu, zlwang@ucsd.edu, yaw048@ucsd.edu, jshang@ucsd.edu}

\newcommand{\MODEL}{LAGER\xspace}
\newcommand{\MODELNearest}{LAGER$_\text{nearest}\:$}
\newcommand{\MODELAngles}{LAGER$_\text{angles}$\:}
\newcommand{\smallsection}[1]{\noindent\textbf{#1.}}

\abstract{
Recent advances of incorporating layout information, typically bounding box coordinates, into pre-trained language models have achieved significant performance in entity recognition from document images. 
Using coordinates can easily model the position of each token, but they are sensitive to manipulations in document images (e.g., shifting, rotation or scaling) which are common in real scenarios.
Such limitation becomes even worse when the training data is limited in few-shot settings.
In this paper, we propose a novel framework, \MODEL, which leverages the topological adjacency relationship among the tokens through learning their relative layout information with graph neural networks. 
Specifically, we consider the tokens in the documents as nodes and formulate the edges based on the topological heuristics.
Such adjacency graphs are invariant to affine transformations, making it robust to the common image manipulations.
We incorporate these graphs into the pre-trained language model by adding graph neural network layers on top of the language model embeddings.
Extensive experiments on two benchmark datasets show that \MODEL significantly outperforms strong baselines under different few-shot settings and also demonstrate better robustness to manipulations.
 \\ \newline  }

\begin{document}

\maketitleabstract

\section{Introduction}

Entity recognition is a fundamental task in document image understanding which aims at identifying and extracting specific segments of text in the document which serve as \textit{header}, \textit{question} or \textit{answer}. However, the named entity recognition in document images is different from the traditional text-only counterparts since document images, such as tables, receipts and forms, involves richer information though the layout structure.
The complex layout and format of these document images provide additional information that can be used to enhance the performance of entity recognition beyond what is possible with only text. Therefore, they present an ideal scenario to use multi-modal techniques. 

Recent existing methods use large self-supervised pre-trained models~\citep{xu2020layoutlm, Xu2020LayoutLMv2MP, huang2022layoutlmv3} for named-entity recognition in document images. These approaches extract the word spans using the standard \texttt{IOBES} tagging schemes~\citep{marquez2005semantic,ratinov2009design} in named entity recognition tasks. The models inherit the architecture from the text-only language models, such as BERT~\cite{devlin2018bert}, RoBERTa~\cite{liu2019roberta}, extend the embedding layer with the layout information, and build layout-aware attention mechanisms. These approaches typically leverage the bounding box coordinates to capture the overall structure of the document, which is straight-forward and has proven to be effective. 
However, we argue that these coordinates-based approaches fail to properly cope with image manipulation, such as shifting, rotation and scaling, which is common in real life. 
These image manipulations make it challenging for coordinate-based approaches to accurately understand the documents, as the coordinates can be significantly altered and the spatial relationships learned by these coordinates are no longer valid.


Given the aforementioned challenges, we propose \MODEL, a \underline{l}ayout-\underline{a}ware \underline{g}raph-based \underline{e}ntity \underline{r}ecognition model. Our new framework further exploits the structural information in these document images utilizing the topological relationship of the entities. We make use of graph neural networks to encode topological relationship in the document. Such practice has been proven effective in other domains such as the web mining from semi-structured web pages~\citep{Lockard2019OpenCeresWO,lockard2020zeroshotceres}, where they build rich representations for text fields on a web page with graphs.
We construct graphs based on the spatial relationship in the document images where the entities correspond to the nodes and the edges are constructed according to heuristics relating to distance and angles between them. In this way, the topological relationship of the entities are explicitly encoded and the resulting graph is robust to the image manipulations mentioned above. We use a Graph Attention Network (GAT) ~\citep{velickovic2018graph} to encode the graph in the latent space and combine it with the rich representations from the pre-trained language models. \MODEL serves as an additional component for the existing layout-aware language models to enhance their robustness to image manipulations and extend their capacity to handle document images with complex layouts.
Our approach is particularly useful in few-shot settings when there is limited data availability for entity recognition, as the graph-based method is efficient to train and easier to generalize. 



\begin{figure*}
 \captionsetup{justification=centering}
    \centering
    \includegraphics[width=0.9\linewidth]{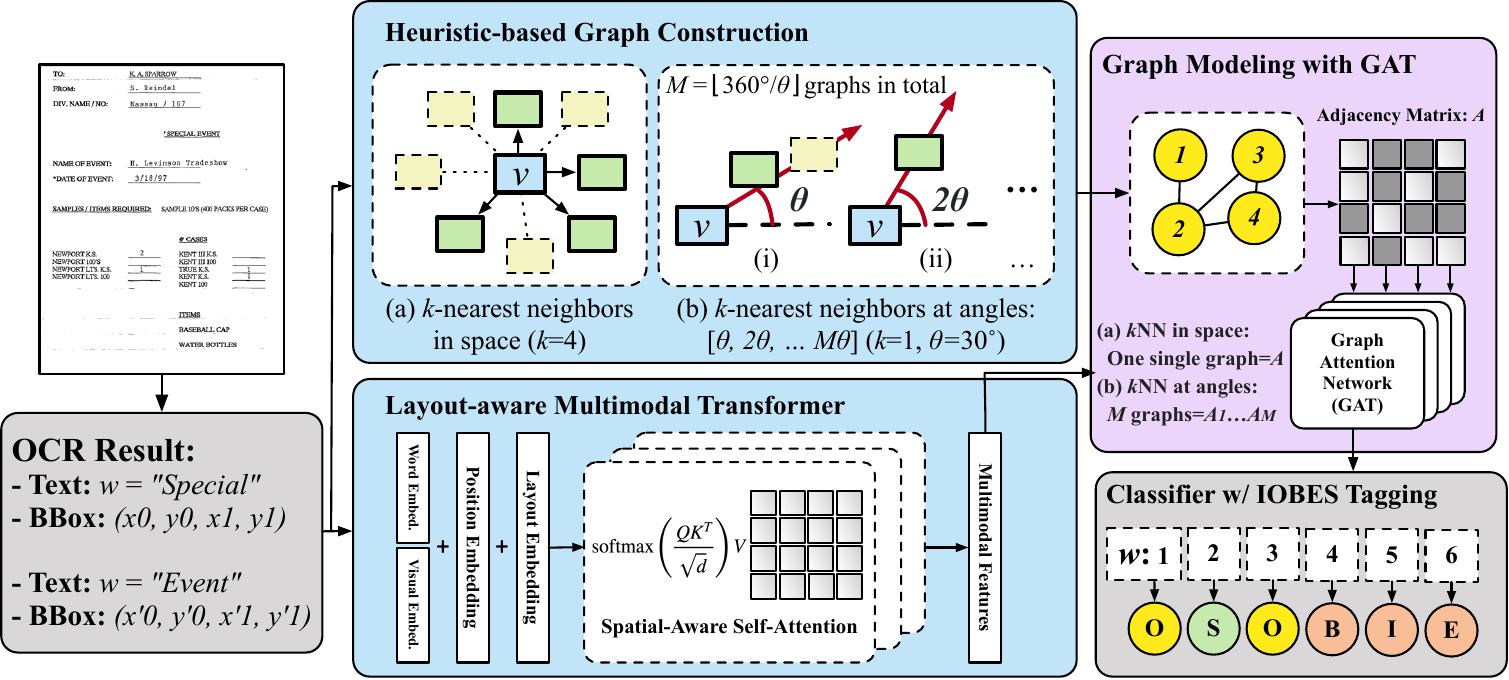}
    \caption{The Framework for \MODEL. Two variants of the model are used based on the heuristic for graph construction. $M$ denotes the number of GATs used. $M=1$ for k-nearest neighbors in space approach. For the the k-nearest neighbors at angles, $M = \lfloor 360 \degree/\theta \rfloor$ and we construct graphs for $[\theta, 2\theta ... M\theta]$.   }

    \label{fig:model}
\end{figure*}

As shown in Figure~\ref{fig:model}, \MODEL extends the architecture of a layout-aware language model which we use as a backbone for our framework~\cite{Xu2020LayoutLMv2MP,huang2022layoutlmv3}. We construct graphs where the nodes correspond to the words in the documents. The edges are constructed based on either k-nearest neighbors of the bounding boxes in space or at multiple angles (the detailed description is given in Section \ref{sec:graph-construction}). The adjacency matrix of this graph along with the hidden states of the backbone language model are given as inputs to the graph attention network (GAT). The enhanced output embeddings from the GAT are then used to perform classification. 

We validate our model using two benchmarks, FUNSD~\citep{jaume2019} and CORD~\citep{park2019cord}. 
Both datasets are from real scenarios and fully-annotated with textual contents and bounding boxes. 
We compare our model with strong baselines and also show how our model is robust to image alterations such as rotations, scaling and shifting. 
We summarize our contribution as follows.
\begin{itemize}[leftmargin=*,nosep]
    \item We propose a novel framework \MODEL that improves existing language models by utilizing the topological relationship of the entities in the document images with Graph Attention Networks.
    \item We show that our approach is robust to image manipulations such as scaling, shifting or rotating, and it is effective to various layout-aware language models.
    \item Extensive experiments on two benchmark datasets and two backbone models demonstrate the effectiveness of \MODEL  under few-shot settings.
\end{itemize}
\noindent\textbf{Reproducibility.} The code and the datasets will be released on Github.

\section{Related Work}
\smallsection{Layout-aware LMs} Given that post-OCR processing has huge potential for various downstream tasks, there are many existing works that have adapted the pre-training in language models such as BERT~\citep{devlin2018bert} to include layout-information. LayoutLM~\citep{xu2020layoutlm} was the first to successfully incorporate layout information in the form of coordinates into the embedding layer of BERT. Following LayoutLM, there was LayoutLMv2~\citep{Xu2020LayoutLMv2MP} which leveraged visual features and improved alignment between words and regions on the page. LayoutLMv3~\citep{huang2022layoutlmv3} like LayoutLMv2 did use visual features but unified text and image masking objectives. There have been other multimodal transformer models such as DocFormer~\citep{docformer2021} which uses text, vision and spatial features. They combine these features using a novel multi-modal self-attention layer. MGDoc~\citep{wang2022mgdoc} aims to exploit the spatial hierarchical relationships between content at different levels of granularity in document images. They do this by encoding page-level, region level, and word-level information at the same time into the pre-training framework. 

\smallsection{Few-shot methods} Recently in the Visually-rich Document Understanding (VrDU) domain, there have been efforts to build robust models under few-shot settings. For semi-structured documents such as business documents, a domain agnostic few-shot learning approach was used~\citep{mandivarapu2021}. Using deep canonical correlation, they were able align the extracted text and image feature vectors. More recently for entity recognition in document images, LASER~\citep{wang2022laser} used a label-aware seq2seq framework. They followed a new labeling scheme that generates the label surface names word-by-word explicitly after generating the entities in few-shot settings.

\smallsection{Graphs in multimodal few-shot settings} Graphs are an extremely useful and general representation of data. This is especially true, when there is some relationship between the data objects in question. Openceres~\citep{Lockard2019OpenCeresWO} for the task of open information extraction from semi structured websites, utilized graphs for their semi-supervised learning approach. 
 ZeroShotCeres~\citep{lockard2020zeroshotceres}, as a successor to OpenCeres used a graph neural network-based approach to build rich representations of text fields on a webpage. They built graphs where each text field became nodes in graphs and used the relationships between the text fields to connect the edges. More recently, FormNet and FormNetv2~\citep{Lee2022FormNetSE,lee2023formnetv2} used graph convolutions to aggregate semantically meaningful information in tokens present in document images.

\section{Methodology}

\subsection{Task Formulation}
Few-shot entity recognition in document images is a subtask of information extraction that seeks to locate and classify named entities into categories using a limited number of training examples. A document image $\mathcal{P}$, consists of textual and layout information. The textual contents correspond to the words, $w$, in the document image and we also have their annotated bounding boxes denoted by $B=(x_0, y_0, x_1, y_1)$ (where $(x_0,y_0)$ and $(x_1,y_1)$ are the top-left and bottom-right corners). These annotations are done by human annotators or OCR engines. These words and bounding boxes are listed sequentially and act as inputs for the textual and layout modalities. The entities are defined as spans of words referring to specific concepts in the document. For example, in FUNSD, the entities correspond to \texttt{question}, \texttt{answer} or \texttt{header}. We train the model with a small subset of training samples (few shots), and test it with full testing set. We denote the the number of training samples as $f$ in $f$-shot training.

\subsection{Pre-trained LM as Backbone}
\label{ssec:backbone}
We perform a thorough literature review and found open source models used in this domain in works such as \citet{wang2022laser}. We find BERT\cite{devlin2018bert}, RoBERTa\cite{liu2019roberta}, LayoutLM\cite{xu2020layoutlm}, LayoutLMv2\cite{Xu2020LayoutLMv2MP} and LayoutLMv3\cite{huang2022layoutlmv3} as models representative for this task. From Table \ref{tab:all_baselines} in the Appendix, we pick the strongest two baselines among these, i.e. LayoutLMv2 and LayoutLMv3. \MODEL is built upon layout aware pre-trained language models such as LayoutLMv2~\citep{Xu2020LayoutLMv2MP} or LayoutLMv3~\citep{huang2022layoutlmv3}. These models are multi-modal transformer models which take text, visual, and layout information as input to incorporate the different interactions. The layout information used are the bounding box coordinates of the tokens in the document.
The output hidden states from the language model is denoted in the form of a feature matrix, $H \in \mathcal{R}^{N\times d}$, 
where $d$ represents the dimension of the hidden state and $N$ denotes the number of tokens in the document.

\subsection{Heuristic-based Graph Construction}\label{sec:graph-construction}
\label{ssec:heuristic}
As discussed previously, the idea of constructing graphs for our document images is to exploit the topological or adjacency relationship present in the entities in the document. Towards this, we construct graphs based on certain heuristics that are used as inputs to the Graph Attention Network (GAT, described in Section \ref{ssec:gat}). \\

\smallsection{Node and Edge definition}  Given a document image page $\mathcal{P}$ with $N$ tokens denoted by $T = \{t_1, t_2, ..., t_N \}$, let $t_i$ refer to the $i$-th token in a text sequence in the dataset. For the token $t_i$, we also know the coordinates of its bounding box, $B_i=(x_{i0},y_{i0},x_{i1},y_{i1})$. Thus for our graph $G = (V, E)$, the vertices $V = \{v_1, v_2, ... , v_N \}$ correspond to all the tokens $T$ and their corresponding bounding boxes. The edges $E$ represent the relationship between pairs of vertices or tokens. We construct an undirected graph, where an edge $e_{ij} \in E$ connects two vertices $v_i$ and $v_j$. Now, we describe how these edges are constructed.  \\

\begin{figure}
    \centering
    \includegraphics[width=0.9\linewidth]{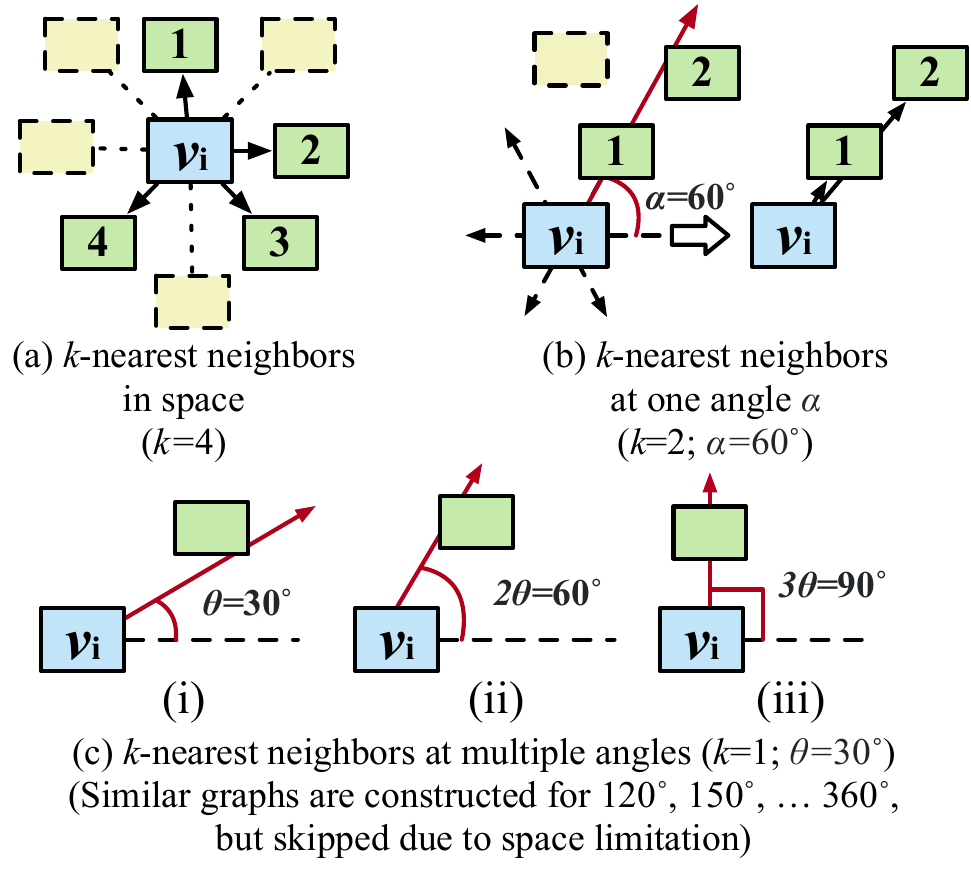}
    \caption{Heuristics for graph construction.
        }
    \label{fig:graph-construction}
    \vspace{-4mm}
\end{figure}

\smallsection{Graph construction} We build graphs based on two heuristics which primarily relate to the Euclidean distance between two tokens in a document. The edges in the graph between tokens are constructed based on either of the following heuristics:
\begin{itemize}[leftmargin=*]
    \item \textbf{k-nearest neighbors in space:} For a token $t_i$, we calculate the Euclidean distance between the corresponding token with all other tokens in $T$. We form edges between $t_i$ and its $k$-closest tokens. If $t_i$ is a $k$-nearest neighbor of $t_j$, there is an edge, or if $t_j$ is a $k$-nearest neighbor of $t_i$, there is an edge. A representative example for $k=4$ is shown in Figure \ref{fig:graph-construction}.a.

    \item \textbf{k-nearest neighbors at multiple angles:} We first describe our method to find the k-nearest neighbor at one angle $\alpha$. Basically, for a token $t_i$, the edges formed are restricted to the k-nearest tokens in the direction of $\alpha$. An example for $k=2$ and $\alpha = 60^{\circ} $ is shown in Figure \ref{fig:graph-construction}.b. 
    \begin{itemize}[leftmargin=*, nosep]
        \item  We draw a ray from the centroid of a token’s bounding box that forms an angle ($\alpha$) with the x-axis (the red ray in Figure \ref{fig:graph-construction}.b).
        \item We find all tokens such that the ray intersects with any part of the token’s bounding box. We then select the $k$-nearest such tokens (the token 1 and 2 in Figure \ref{fig:graph-construction}.b).
    \end{itemize}
    The approach for one single angle collects the information in that particular direction. We create multiple graphs to represent the global topological relation of each token. We pick an angle $\theta$ and $M = \lfloor360\degree/\theta\rfloor$ graphs are created with $\alpha\in\{\theta,2\theta,...\}$ (the graphs in Figure \ref{fig:graph-construction}.c).
    

\end{itemize}
After constructing the graph(s), we would create one adjacency matrix $A$ for k-nearest neighbors in space or multiple adjacency matrices $A_1,...,A_M$ for k-nearest neighbors at multiple angles. For simplicity, we denote these adjacency matrices by $A \in \mathcal{R}^{N\times N}$ to represent the topological structure when there is no ambiguity. And $A_{v_i, v_j}=1$ if and only if an edge $e = (v_i, v_j)$ exists in our graph. 

We believe that constructing the graph using the heuristics described above allows us to capture some relationships between tokens in the document that is not leveraged by using just the layout aware pre-trained language model. The graphs, especially the one constructed using k-nearest neighbors at multiple angles can preserve the topological relationship. It also helps in recovering the relative positions of the different bounding boxes even in cases described in Section \ref{ssec:image_manip} where the documents are altered with scaling, rotations or shifting. 


\subsection{Graph Modeling with GAT}
\label{ssec:gat}
Our model combines the representations from the pre-trained language model with the graph we construct for a document described in the previous section. For this, we use a graph neural network, specifically Graph Attention Network (GAT)~\citep{velickovic2018graph} which is a commonly-used graph neural network architecture and has shown state-of-the-art performance on various tasks. The GAT computes latent representations of each node in the graph, by attending over its neighbors following a self-attention strategy. To stabilise the learning process of self-attention, the graph attention layer uses multi-head attention as in the Transformer architecture~\citep{vaswani2017attn}. Namely, the operations of this layer are independently replicated $h$ times (each with different parameters), and outputs are feature-wise concatenated. 
The inputs to the GAT are, a feature matrix $H$ and an adjacency matrix $A$. We obtain the adjacency matrix based on the graph construction explained in Section \ref{ssec:heuristic}. We obtain the feature matrix $H$ from the output of the backbone language model as described in Section \ref{ssec:backbone}. Thus, each node in the graph (token in a document) contains a corresponding embedding. We get an enhanced output representation, $H' = GAT (H, A)$. 


We use two variants of our model \MODELNearest and \MODELAngles based on the two heuristics of graph construction described in Section \ref{ssec:heuristic}.
\begin{itemize}[leftmargin=*]
\item \textbf{k-nearest neighbors in space:} For this, we use a single GAT ($M=1$) whose adjacency matrix is based on the k-nearest neighbors in space heuristic.

\item  \textbf{k-nearest neighbors at multiple angles:} Based on this heuristic, we construct multiple graphs to gather the spatial information around the token. That is, we construct $M$ graphs that evenly distribute in the space where $M = \lfloor 360 \degree/\theta \rfloor$. For example, if $\theta = 60\degree$, then we construct 6 graphs for $0\degree, 60\degree, 120\degree, 180\degree, 240\degree, 300\degree$. 
For each of these $M$ graphs, we use a specific GAT (each with their respective parameters) and then take an average of all the GAT outputs. Specifically,

\begin{align*}
    H^\prime_{1} = GAT_1 (H, A_1) &... H^\prime_{M} = GAT_M (H, A_M) \\
    H^\prime = &\frac{1}{M}\sum H^\prime_i
\end{align*}
where $A_i$ is the adjacency matrix constructed with $\frac{i-1}{M} \cdot \lfloor \frac{360\degree}{\theta} \rfloor$.
\end{itemize}

Once we have the embeddings $H'$ from the GAT, it undergoes a linear affine transformation which is represented by the classifier layer in Figure \ref{fig:model}. Following this, the model predicts the $\{I, O, B, E, S\}$ tags for each token in the document and uses sequence labeling to detect each type of entity for the corresponding dataset.


\begin{figure}[t] 
\centering
  \subfigure[Rotation with $\delta = 8\degree$]{%
    \includegraphics[height=1.5cm, width=0.8\linewidth]{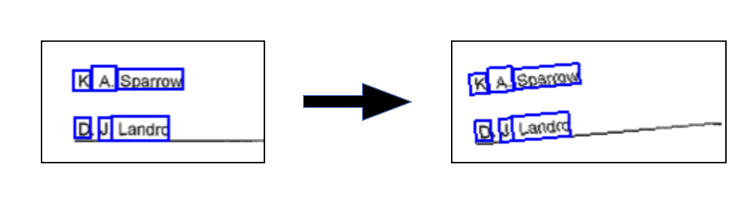}

  } 
  \subfigure[Scaling by a factor of 4 ($s_w=2, s_h=2$)]{%
    \includegraphics[height=3cm, width=0.8\linewidth ]{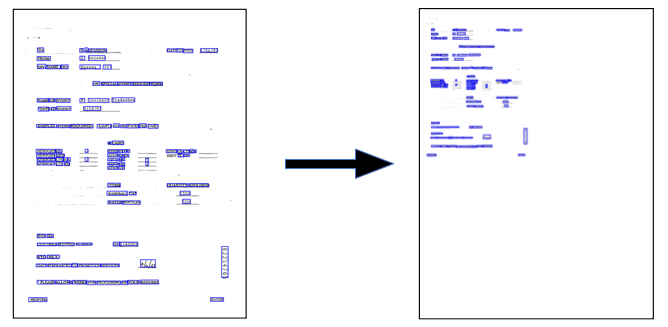}

  } 
  \hfill 
  \subfigure[Shifting with $a=10$]{%
    \includegraphics[height=1.6cm,width=0.8\linewidth]{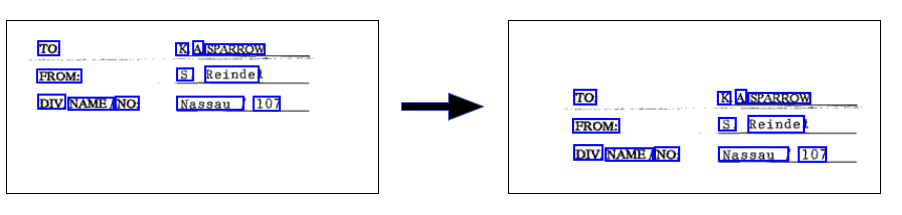}
  } 
  \caption{Representative examples of the image manipulations} 
  \label{fig:image_manip}
\end{figure}
\vspace{-2mm}
\subsection{Image manipulations}
\label{ssec:image_manip}
In real-world scenarios, document images are often not in ideal, regular conditions and can have some alterations such as shifting, rotation or scaling.
As illustrated in Figure \ref{fig:image_manip}, we perform three types of manipulations to the document images: 

\begin{itemize}[leftmargin=*]
    \item \textbf{Shifting:} In every document, for each bounding box $B=(x_0, y_0, x_1, y_1)$, we translate the coordinates of the four corners with a translation vector $(a,a)$. Thus, the modified bounding box now is $B'=(x_0+a, y_0+a, x_1+a, y_1+a)$. 
 
    \item \textbf{Rotation:} We rotate each document by a small angle $\delta$ about the bottom left corner of the document. Thus, for each bounding box $B=(x_0, y_0, x_1, y_1)$, we rotate the bounding box around $(x_0, y_1)$, i.e. the bottom left corner of the bounding box is the center of rotation. Thus, for every $(x, y)$ corner of a bounding box, we have the resulting rotated point $(x', y')$, where
    \vspace{-2pt}
    \begin{align*}
    \vspace{-2pt}
        \text{$x' = (x - x_0) \cdot \cos(\delta) - (y - y_1) \cdot \sin(\delta) + x_0$}\\
        \text{$y' = (x - x_0) \cdot \sin(\delta) + (y - y_1) \cdot \cos(\delta) + y_1$}
    \end{align*}

    \vspace{-5pt}
    \item \textbf{Scaling:} We scale down, i.e. reduce the size of the entire document by a factor. If $w$, $h$ denote the width and height of the document and $s_w$, $s_h$ denote the factor of scaling for the width and height. In every document, for each bounding box $B=(x_0, y_0, x_1, y_1)$, the scaled down coordinates would now be $B=(x_0/s_w, y_0/s_h, x_1/s_w, y_1/s_h)$.
\end{itemize}

\begin{table}[t]
    \centering
    \resizebox{\linewidth}{!}{
    \small
    \setlength{\tabcolsep}{0.8mm}{
\begin{tabular}{lccc}
\toprule
\textbf{Dataset} & \textbf{\# Train Pages} & \textbf{\# Test Pages} & \textbf{\# Entities / Page} \\
\midrule
FUNSD & 149   & 50    & 42.86 \\
CORD & 800   & 100   & 13.82 \\
\bottomrule
\end{tabular}%
}}
    \caption{Dataset Statistics.}
    \label{tab:dataset}
    
\end{table}%

\section{Experiments}
We conduct extensive experiments on the FUNSD~\citep{jaume2019} and the CORD~\citep{park2019cord} datasets under few-shot settings. We also look at how the vanilla baseline models and our proposed models fare under environments where the document images have been manipulated. We also look at some example case studies from both the datasets.

All the experiments under the few-shot settings using few-shot sizes ranging from 1 to 10. We use 6 different random seeds to select the few-shot samples from our training set. We train the different models for a particular few-shot size using the same data and compute the average performance and the standard deviation across the 6 seeds. We report only the result of 2, 3, 4, 5 and 6 shots due to space limitation. We report the results for all 10 few-shot sizes in the Appendix. For model evaluation, the results are first converted into \texttt{IOBES} tagging style and we then compute the word-level precision, recall and F-1 score using the seqeval APIs~\citep{seqeval}. All implementation details including hyperparameters used for all the experiments is in the Appendix \ref{ssec:impl}.

\subsection{Datasets}
Our experiments are conducted on two real-world data collections: FUNSD and CORD. Both datasets provide rich annotations for the document images and include the words and the word-level bounding boxes. The details and statistics (Table \ref{tab:dataset}) of these two datasets are as follows.
\begin{itemize}[nosep, leftmargin=*]
    \item \textbf{FUNSD:} FUNSD consists of 199 fully-annotated, noisy-scanned forms with various appearance and format which makes the form understanding task more challenging. The word spans in this datasets are annotated with three different labels: \texttt{header}, \texttt{question} and \texttt{answer}, and the rest words are annotated as \texttt{other}. We use the original label names.
    \item \textbf{CORD:} CORD consists of about 1000 receipts with annotations of bounding boxes and textual contents. The word spans in this datasets are annotated with 30 different labels. The broad categories include \texttt{menu}, \texttt{sub-total} and \texttt{total}. We use the original label names.
\end{itemize}


\begin{table*}[t]
\centering

\resizebox{\linewidth}{!}{
\setlength{\tabcolsep}{3mm}{
\begin{tabular}{ccccccccc}
\toprule
\multirow{2}{*}{} & \multirow{2}{*}{$|\mathcal{P}|$} & \multirow{2}{*}{\textbf{Model}} & \multicolumn{3}{c}{\textbf{FUNSD}} & \multicolumn{3}{c}{\textbf{CORD}} \\ \cline{4-9} 
 &  &  & \textbf{Precision} & \textbf{Recall} & \textbf{F-1} & \textbf{Precision} & \textbf{Recall} & \textbf{F-1} \\ \hline
\multirow{15}{*}{\rotatebox{90}{LayoutLMv2}} & \multirow{3}{*}{2} & Vanilla & 38.61$\pm$4.04 & 53.8$\pm$8.35 & 44.82$\pm$5.3 & 43.34$\pm$2.07 & 55.85$\pm$2.43 & 48.66$\pm$2.13 \\
 &  & + \MODELNearest & 41.4$\pm$3.26 & 52.08$\pm$9.89 & 45.74$\pm$5.32 & 43.42$\pm$2.45 & 55.89$\pm$3.15 & 48.87$\pm$2.73 \\
 &  & + \MODELAngles & \textbf{41.45$\pm$2.22} & \textbf{54.21$\pm$2.99} & \textbf{46.9$\pm$1.66} & \textbf{43.77$\pm$2.77} & \textbf{56.22$\pm$2.94} & \textbf{49.21$\pm$2.8} \\ \cline{2-9} 
 & \multirow{3}{*}{3} & Vanilla & 47.73$\pm$2.91 & 61.16$\pm$7.02 & 53.55$\pm$4.36 & 50.52$\pm$2.87 & 61.03$\pm$3.8 & 54.69$\pm$3.19 \\
 &  & + \MODELNearest & \textbf{48.72$\pm$3.33} & \textbf{63.12$\pm$8.69} & \textbf{54.9$\pm$5.42} & \textbf{50.81$\pm$3.03} & \textbf{61.49$\pm$3.48} & \textbf{55.02$\pm$3.09} \\
 &  & + \MODELAngles & 48.48$\pm$3.46 & 60.68$\pm$9.14 & 53.79$\pm$5.73 & 50.15$\pm$2.79 & 61.28$\pm$2.98 & 54.53$\pm$2.7 \\ \cline{2-9} 
 & \multirow{3}{*}{4} & Vanilla & 51.13$\pm$2.16 & 64.57$\pm$6.8 & 56.95$\pm$3.7 & 55.18$\pm$2.25 & 65.96$\pm$3.35 & 59.88$\pm$2.66 \\
 &  & + \MODELNearest & \textbf{52.59$\pm$3.71} & \textbf{67.23$\pm$4.75} & \textbf{58.93$\pm$3.55} & 54.86$\pm$2.57 & 65.87$\pm$3.88 & 59.85$\pm$3.05 \\
 &  & + \MODELAngles & 50.86$\pm$3.43 & 66.47$\pm$4.56 & 57.59$\pm$3.63 & \textbf{55.52$\pm$2.56} & \textbf{66.57$\pm$3.45} & \textbf{60.54$\pm$2.89} \\ \cline{2-9} 
 & \multirow{3}{*}{5} & Vanilla & 53.46$\pm$2.43 & 64.57$\pm$6.71 & 58.36$\pm$3.63 & 57.59$\pm$3.11 & 68.11$\pm$3.24 & 62.27$\pm$3.06 \\
 &  & + \MODELNearest & \textbf{54.0$\pm$2.54} & \textbf{67.77$\pm$4.98} & \textbf{60.02$\pm$2.83} & 57.49$\pm$3.1 & 68.27$\pm$2.55 & 62.41$\pm$2.82 \\
 &  & + \MODELAngles & 52.98$\pm$3.25 & 67.23$\pm$5.08 & 59.19$\pm$3.53 & \textbf{57.88$\pm$2.73} & \textbf{68.27$\pm$2.48} & \textbf{62.63$\pm$2.52} \\ \cline{2-9} 
 & \multirow{3}{*}{6} & Vanilla & 56.92$\pm$1.59 & 67.59$\pm$2.9 & 61.77$\pm$1.77 & 60.28$\pm$2.47 & 70.01$\pm$2.48 & 64.41$\pm$2.43 \\
 &  & + \MODELNearest & 57.45$\pm$2.57 & 71.14$\pm$2.83 & 63.53$\pm$2.25 & \textbf{60.63$\pm$2.88} & \textbf{70.49$\pm$2.69} & \textbf{65.19$\pm$2.78} \\
 &  & + \MODELAngles & \textbf{58.14$\pm$3.79} & \textbf{71.34$\pm$3.02} & \textbf{63.95$\pm$2.21} & 59.97$\pm$2.37 & 70.04$\pm$2.28 & 64.6$\pm$2.26 \\ \bottomrule
\end{tabular}
}}
\caption{Evaluation results with LayoutLMv2 as the backbone model on different few-shot sizes. \textbf{Bold} denotes the best model}
\label{tab:lmv2-results}
\end{table*}
\begin{table*}[t]
\centering

\resizebox{\linewidth}{!}{
\setlength{\tabcolsep}{3mm}{
\begin{tabular}{ccccccccc}
\toprule
\multirow{2}{*}{} & \multirow{2}{*}{$|\mathcal{P}|$} & \multirow{2}{*}{\textbf{Model}} & \multicolumn{3}{c}{\textbf{FUNSD}} & \multicolumn{3}{c}{\textbf{CORD}} \\ \cline{4-9} 
 &  &  & \textbf{Precision} & \textbf{Recall} & \textbf{F-1} & \textbf{Precision} & \textbf{Recall} & \textbf{F-1} \\ \hline
\multirow{15}{*}{\rotatebox{90}{LayoutLMv3}} & \multirow{3}{*}{2} & Vanilla & 44.29$\pm$6.14 & 58.96$\pm$7.2 & 50.43$\pm$6.03 & 47.21$\pm$6.25 & 58.99$\pm$4.94 & 52.41$\pm$5.85 \\
 &  & + \MODELNearest & \textbf{49.82$\pm$6.06} & \textbf{59.55$\pm$8.91} & \textbf{54.09$\pm$6.54} & \textbf{48.68$\pm$5.72} & 60.19$\pm$4.23 & \textbf{53.79$\pm$5.24} \\
 &  & + \MODELAngles & 46.8$\pm$6.46 & 58.15$\pm$8.92 & 51.61$\pm$6.42 & 48.15$\pm$5.07 & \textbf{60.3$\pm$3.57} & 53.51$\pm$4.58 \\ \cline{2-9} 
 & \multirow{3}{*}{3} & Vanilla & 59.66$\pm$4.92 & 72.2$\pm$7.65 & 65.29$\pm$5.92 & 51.34$\pm$6.55 & 62.49$\pm$5.47 & 56.34$\pm$6.2 \\
 &  & + \MODELNearest & 62.18$\pm$5.13 & \textbf{73.12$\pm$7.3} & \textbf{67.12$\pm$5.56} & \textbf{53.08$\pm$7.32} & \textbf{64.3$\pm$5.55} & \textbf{58.1$\pm$6.73} \\
 &  & + \MODELAngles & \textbf{60.73$\pm$5.09} & 72.41$\pm$7.64 & 65.97$\pm$5.71 & 52.77$\pm$7.17 & 63.72$\pm$5.55 & 57.68$\pm$6.63 \\ \cline{2-9} 
 & \multirow{3}{*}{4} & Vanilla & 65.32$\pm$3.89 & 77.97$\pm$2.26 & 71.06$\pm$3.04 & 54.18$\pm$5.01 & 64.92$\pm$3.76 & 59.04$\pm$4.53 \\
 &  & + \MODELNearest & \textbf{67.86$\pm$3.3} & \textbf{78.73$\pm$2.57} & \textbf{72.86$\pm$2.69} & \textbf{56.28$\pm$4.24} & \textbf{66.47$\pm$3.29} & \textbf{60.94$\pm$3.86} \\
 &  & + \MODELAngles & 65.93$\pm$3.28 & 77.22$\pm$3.45 & 71.08$\pm$2.81 & 55.38$\pm$4.63 & 65.99$\pm$3.79 & 60.21$\pm$4.3 \\ \cline{2-9} 
 & \multirow{3}{*}{5} & Vanilla & 67.14$\pm$5.17 & 77.88$\pm$2.62 & 72.07$\pm$4.01 & 58.55$\pm$2.82 & 67.03$\pm$2.46 & 62.49$\pm$2.57 \\
 &  & + \MODELNearest & 69.6$\pm$2.57 & 79.72$\pm$1.66 & 74.3$\pm$1.94 & \textbf{59.84$\pm$3.27} & 68.36$\pm$2.34 & \textbf{63.8$\pm$2.76} \\
 &  & + \MODELAngles & \textbf{70.32$\pm$1.41} & \textbf{80.86$\pm$1.23} & \textbf{75.22$\pm$1.1} & 59.37$\pm$4.09 & \textbf{68.48$\pm$3.08} & 63.58$\pm$3.63 \\ \cline{2-9} 
 & \multirow{3}{*}{6} & Vanilla & 71.19$\pm$3.75 & 80.83$\pm$1.09 & 75.68$\pm$2.58 & 60.91$\pm$3.51 & 69.16$\pm$2.76 & 64.76$\pm$3.16 \\
 &  & + \MODELNearest & 72.71$\pm$3.42 & 81.53$\pm$1.98 & \textbf{76.84$\pm$2.58} & \textbf{61.8$\pm$5.14} & 70.0$\pm$3.75 & 65.63$\pm$4.53 \\
 &  & + \MODELAngles & \textbf{72.31$\pm$3.7} & \textbf{81.65$\pm$1.81} & 76.67$\pm$2.7 & 61.56$\pm$4.49 & \textbf{70.3$\pm$3.33} & \textbf{65.63$\pm$3.98} \\ \bottomrule
\end{tabular}
}}
\caption{Evaluation results with LayoutLMv3 as the backbone model on different few-shot sizes. \textbf{Bold} denotes the best model}
\label{tab:lmv3-results}
\vspace{-3mm}
\end{table*}

\subsection{Baselines}
\label{ssec:baselines}
 Based on Table \ref{tab:all_baselines} in the Appendix, we select the two strongest baselines that are representative for our task, i.e. LayoutLMv2 and LayoutLMv3. In our model \MODEL, we use LayoutLMv2~\citep{Xu2020LayoutLMv2MP} and LayoutLMv3~\citep{huang2022layoutlmv3} as backbones . We evaluate \MODEL against vanilla LayoutLMv2 and LayoutLMv3 in few-shot setting. 

\begin{itemize}[nosep, leftmargin=*]
\item \textbf{LayoutLMv2:}  is a multi-modal language model which is an improved version of LayoutLM~\citep{xu2020layoutlm}. It integrates the visual information in the pre-training stage to learn the cross-modality interaction between visual and textual information. 
\item \textbf{LayoutLMv3: } is another large multi-modal language model which aims to mitigate the discrepancy between text and image modalities in other models such as LayoutLM and LayoutLMv2. It facilitates multimodal representation learning by unifying the text and image masking.
\end{itemize}
For all our experiments we use the base version of these models and follow the \texttt{IOBES} tagging scheme.
\begin{table*}[t]

\centering
\resizebox{\linewidth}{!}{
  \setlength{\tabcolsep}{1mm}{
\begin{tabular}{lcccccc|cccc|cccc}
\toprule
\multirow{3}{*}{} & \multirow{3}{*}{$|\mathcal{P}|$} & \multirow{3}{*}{Model} & \multicolumn{4}{c|}{Shift ($a = 20$)} & \multicolumn{4}{c|}{Scale ($s_w = 2$, $s_h = 2$)} & \multicolumn{4}{c}{Rotation ($\delta = 8\degree$)} \\ \cline{4-15} 
 &  &  & \multicolumn{2}{c}{FUNSD} & \multicolumn{2}{c|}{CORD} & \multicolumn{2}{c}{FUNSD} & \multicolumn{2}{c|}{CORD} & \multicolumn{2}{c}{FUNSD} & \multicolumn{2}{c}{CORD} \\ \cline{4-15} 
 &  &  & \textbf{F-1} & \textbf{Diff.} & \textbf{F-1} & \textbf{Diff.} & \textbf{F-1} & \textbf{Diff.} & \textbf{F-1} & \textbf{Diff.} & \textbf{F-1} & \textbf{Diff.} & \textbf{F-1} & \textbf{Diff.} \\ \hline
\multirow{15}{*}{\rotatebox{90}{LayoutLMv3}} & \multirow{3}{*}{2} & Vanilla & 49.28$\pm$5.69 & 1.15 & 50.53$\pm$5.58 & 1.88 & 32.66$\pm$15.64 & 17.77 & 38.77$\pm$6.62 & 13.64 & 48.11$\pm$5.77 & 2.33 & 48.39$\pm$5.31 & 4.02 \\
 &  & + \MODELNearest & \textbf{53.31$\pm$5.03} & 0.78 & 51.97$\pm$5.24 & 1.82 & \textbf{38.07$\pm$16.16} & 16.02 & \textbf{40.66$\pm$7.63} & \textbf{13.13} & \textbf{52.56$\pm$6.22} & \textbf{1.53} & 50.4$\pm$5.48 & 3.39 \\
 &  & + \MODELAngles & 51.32$\pm$5.83 & \textbf{0.29} & \textbf{51.98$\pm$4.45} & \textbf{1.53} & 36.14$\pm$15.32 & \textbf{15.47} & 39.96$\pm$7.98 & 13.55 & 49.58$\pm$6.64 & 2.03 & \textbf{50.3$\pm$4.97} & \textbf{3.21} \\ \cline{2-15} 
 & \multirow{3}{*}{3} & Vanilla & 63.22$\pm$5.5 & 2.07 & 54.15$\pm$5.76 & 2.19 & 46.44$\pm$15.39 & 18.85 & 43.04$\pm$7.23 & 13.3 & 63.24$\pm$5.69 & 2.05 & 51.2$\pm$5.72 & 5.14 \\
 &  & + \MODELNearest & \textbf{65.34$\pm$4.58} & \textbf{1.78} & \textbf{55.94$\pm$6.22} & 2.16 & 48.78$\pm$12.16 & 18.34 & 44.6$\pm$8.52 & 13.5 & \textbf{65.36$\pm$5.46} & \textbf{1.76} & \textbf{52.84$\pm$6.71} & 5.25 \\
 &  & + \MODELAngles & 63.98$\pm$4.37 & 1.99 & 55.72$\pm$5.96 & \textbf{1.96} & \textbf{49.61$\pm$11.33} & \textbf{16.36} & \textbf{44.63$\pm$7.46} & \textbf{13.05} & 64.04$\pm$5.65 & 1.92 & 52.72$\pm$6.56 & \textbf{4.96} \\ \cline{2-15} 
 & \multirow{3}{*}{4} & Vanilla & 68.66$\pm$3.13 & 2.4 & 56.8$\pm$3.84 & 2.24 & 52.7$\pm$9.16 & 18.36 & 44.17$\pm$5.89 & 14.47 & 68.84$\pm$3.34 & \textbf{2.22} & 54.57$\pm$4.24 & 4.47 \\
 &  & + \MODELNearest & \textbf{70.62$\pm$2.81} & 2.24 & \textbf{58.97$\pm$3.35} & \textbf{1.97} & \textbf{54.86$\pm$9.4} & \textbf{18} & \textbf{46.64$\pm$6.54} & \textbf{14.3} & \textbf{70.4$\pm$3.06} & 2.46 & \textbf{56.9$\pm$3.17} & \textbf{4.04} \\
 &  & + \MODELAngles & 68.92$\pm$2.66 & \textbf{2.16} & 58.06$\pm$3.87 & 2.15 & 52.99$\pm$11.59 & 18.09 & 45.86$\pm$6.12 & 14.35 & 68.71$\pm$3.0 & 2.36 & 55.76$\pm$3.63 & 4.45 \\ \cline{2-15} 
 & \multirow{3}{*}{5} & Vanilla & 70.01$\pm$4.08 & 2.06 & 59.47$\pm$3.23 & 3.02 & 47.18$\pm$15.26 & 24.89 & 45.02$\pm$4.41 & 17.47 & 69.62$\pm$4.13 & 2.45 & 57.31$\pm$2.85 & 5.18 \\
 &  & + \MODELNearest & 72.43$\pm$1.66 & \textbf{1.87} & \textbf{61.73$\pm$2.52} & \textbf{2.07} & \textbf{57.18$\pm$8.11} & \textbf{17.12} & \textbf{49.83$\pm$3.18} & \textbf{13.97} & 72.36$\pm$2.66 & \textbf{1.93} & \textbf{58.88$\pm$2.33} & \textbf{4.91} \\
 &  & + \MODELAngles & \textbf{73.28$\pm$1.15} & 1.94 & 61.01$\pm$3.15 & 2.57 & 56.73$\pm$5.68 & 18.49 & 48.55$\pm$3.08 & 15.03 & \textbf{72.86$\pm$2.1} & 2.35 & 58.33$\pm$2.54 & 5.25 \\ \cline{2-15} 
 & \multirow{3}{*}{6} & Vanilla & 73.58$\pm$1.43 & 2.1 & 61.91$\pm$3.18 & 2.85 & 53.07$\pm$13.76 & 22.61 & 47.18$\pm$1.7 & 17.58 & 73.54$\pm$3.15 & 2.14 & 60.19$\pm$2.04 & 4.57 \\
 &  & + \MODELNearest & 74.59$\pm$2.79 & 2.25 & \textbf{63.35$\pm$4.41} & \textbf{2.28} & \textbf{57.0$\pm$11.24} & \textbf{19.84} & \textbf{52.34$\pm$3.02} & \textbf{13.29} & 75.09$\pm$2.75 & \textbf{1.75} & \textbf{61.26$\pm$3.25} & \textbf{4.37} \\
 &  & + \MODELAngles & \textbf{75.07$\pm$2.52} & \textbf{1.6} & 63.24$\pm$3.64 & 2.39 & 56.53$\pm$12.75 & 20.14 & 51.68$\pm$4.05 & 13.95 & \textbf{74.72$\pm$2.94} & 1.95 & 60.93$\pm$2.67 & 4.69 \\ \bottomrule
\end{tabular}

}
\vspace{-3mm}
}
\caption{Evaluation results on image manipulation with shifting ($a = 20$), scaling by a factor of $4$ ($s_w=2$,$s_h=2$) and rotation with $\delta = 8\degree$ with LayoutLMv3 as backbone. The column Diff. refers to the difference in F1-scores between results in setting without manipulation (Table \ref{tab:lmv3-results}) and with manipulation. \textbf{Bold} denotes the best model}
\label{tab:image-manip}
\end{table*}
\subsection{Implementation Details}
\label{ssec:impl}
We build our model on top of LayoutLMv2/LayoutLMv3 as our backbone language model. We use the Transformers~\citep{wolf2019huggingface} and also utilize the repository of \citet{unilm} to build our model. We use one NVIDIA A6000 to finetune with batch size of 8. We optimize the model with AdamW optimizer and the learning rate is $5\times10^{-5}$.

We ran extensive experiments for various intuitive choices of hyperparameters. For the value of $k$ during graph construction, we try different values like 1,2,4, and 8. All results reported for both heuristics use $k=4$.  For the $k$-nearest neighbors at multiple angles, the idea is to capture the topological relationship of a token. Thus, it's quite natural to divide the 2D plane into 2,4,6 or 8 halves, i.e. angles such as 90, 60, 45, 30, 15, etc. were tried. Though most of these choices work great, the one in the experiments reported use $\theta = 60^\circ$. Thus, we use $M = 360\degree/60\degree = 6$ GATs and average the outputs of these different GATs when we run the experiments. For all our experiments, we set the number of heads in the GAT, to $h=4$. 

\subsection{Few-shot Experimental results}

We report our results using the two baseline models described in Section \ref{ssec:baselines} in Tables \ref{tab:lmv2-results} and \ref{tab:lmv3-results}. For the baseline and as the backbone language model in \MODEL, Table \ref{tab:lmv2-results} and \ref{tab:lmv3-results} use LayoutLMv2 and LayoutLMv3 respectively.
The results are reported on two versions of our model, \MODELNearest and \MODELAngles for the two heuristics of graph construction described in Section \ref{ssec:heuristic} and \ref{ssec:gat}. We observe that our model achieves significant performance improvements compared with the baselines for both FUNSD and CORD datasets. We see in Table \ref{tab:lmv2-results}, there is on average relative improvements of $4\%$ and $1.5\%$ in terms of F-1 score for FUNSD and CORD respectively over the vanilla LayoutLMv2 baseline. For Table \ref{tab:lmv3-results}, we see an average relative improvement in terms of F-1 score by $4\%$ and $3\%$ for FUNSD and CORD respectively over the LayoutLMv3 baseline. We see similar gains in performance for precision and recall in both the tables. 

We also analyze the filewise results of the test set instances for both FUNSD and CORD. That is, for each individual test set instance, we compare the filewise F-1 scores of our models with the baseline. We observe that when using LayoutLMv2 as the backbone, our models on average improve over the baseline for $58\%$ and $62\%$ of our test set instances for FUNSD and CORD respectively. Similarly, when using LayoutLMv3 as the backbone, our models on average improve over the baseline for $65\%$ and $67\%$ of our test set instances for FUNSD and CORD respectively. This shows that \MODELNearest and \MODELAngles provide more confident predictions leading to the overall performance improvement for the entity recognition task.

Based on these comparisons, we conclude that our proposed framework is superior to the traditional vanilla language model baselines in few-shot settings.

\begin{figure*}[t] 
  \centering
  
  \subfigure[Ground truth]{%
    \includegraphics[height=1.75cm, width=0.45\linewidth]{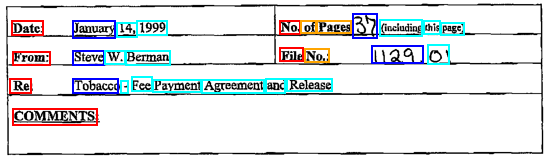}
    \vspace{-5mm}
  } 
  \hfill 
  \subfigure[LayoutLMv2]{%
    \includegraphics[height=1.75cm, width=0.45\linewidth]{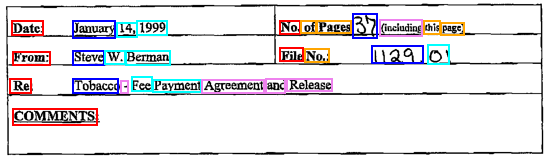}
    \vspace{-5mm}
  } 
  \hfill 
  \subfigure[LayoutLMv2 + \MODELNearest]{%
    \includegraphics[height=1.75cm,width=0.45\linewidth]{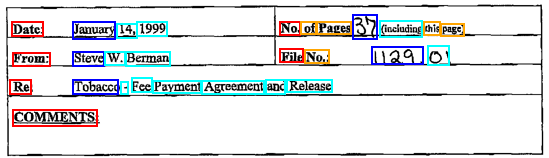}
    \vspace{-5mm}
  } 
  \hfill 
  \subfigure[LayoutLMv2 + \MODELAngles]{%
    \includegraphics[height=1.75cm,width=0.45\linewidth]{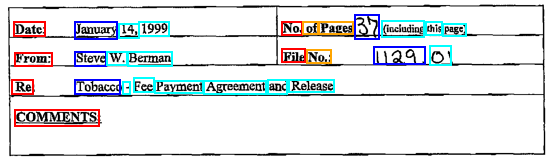}
    \vspace{-5mm}
  } 
  \vspace{-3mm}
  \caption{Case studies from FUNSD. \colorbox[RGB]{0,0,255}{\textcolor[RGB]{0,0,255}{Bl}},
  \colorbox[RGB]{0,255,255}{\textcolor[RGB]{0,255,255}{Bl}},
  \colorbox[RGB]{255,0,0}{\textcolor[RGB]{255,0,0}{Bl}},
  \colorbox[RGB]{255,165,0}{\textcolor[RGB]{255,165,0}{Bl}},
  \colorbox[RGB]{238,130,238}{\textcolor[RGB]{238,130,238}{Bl}} denote \texttt{B-ANSWER}, \texttt{I-ANSWER}, \texttt{B-QUESTION}, \texttt{I-QUESTION} and \texttt{OTHER} respectively. 
   }
  \label{fig:lmv2-cases}
  \vspace{-3mm}
\end{figure*}

\begin{figure*}[t] 
\centering
  \subfigure[Ground truth]{%
    \includegraphics[height=1cm, width=0.22\linewidth]{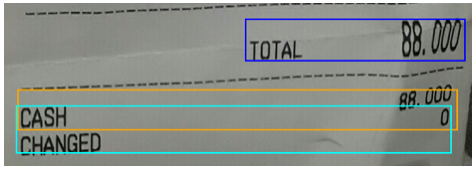}
    \vspace{-5mm}
  } 
  \hfill 
  \subfigure[LayoutLMv3]{%
    \includegraphics[height=1cm, width=0.22\linewidth]{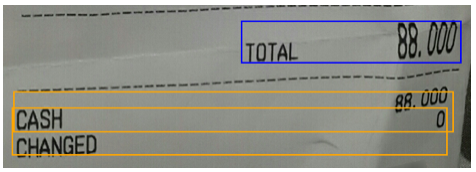}
    \vspace{-5mm}
  } 
    \hfill 
  \subfigure[LayoutLMv3 + \MODELNearest ]{%
    \includegraphics[height=1cm, width=0.22\linewidth]{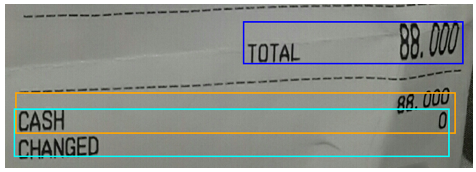}
    \vspace{-5mm}
  } 
  \hfill 
  \subfigure[LayoutLMv3 + \MODELAngles]{%
    \includegraphics[height=1cm, width=0.22\linewidth]{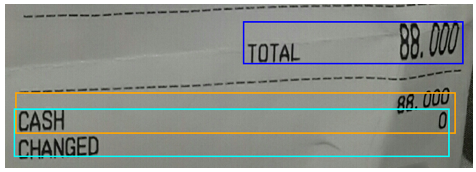}
    \vspace{-5mm}
  } 
  \vspace{-3mm}
  \caption{Case studies from CORD. \colorbox[RGB]{0,0,255}{\textcolor[RGB]{0,0,255}{Bl}},  
  \colorbox[RGB]{255, 165, 0}{\textcolor[RGB]{255, 165, 0}{Bl}},
  \colorbox[RGB]{0, 255, 255}{\textcolor[RGB]{0, 255, 255}{Bl}},
  denote the tags for \texttt{TOTAL.TOTAL\_PRICE} \texttt{TOTAL.CASHPRICE} and \texttt{TOTAL.CHANGEPRICE} respectively.
  } 
  \vspace{-3mm}
  \label{fig:lmv3-cases}
\end{figure*}

\subsection{Experiments with image manipulation}

In these experiments, for the models in Table \ref{tab:lmv3-results}  we manipulate the test-set images during inference as described in Section \ref{ssec:image_manip}. We perform experiments with various factors of shifting, scaling and rotation and observe similar evaluation results. We show an instance of each here due to space constraints.
 We use shifting with a factor of $a=20$, scaling with a factor of $4$, i.e $s_h=2, s_w =2$ and rotation with $\delta = 8\degree$. The results are reported in Table \ref{tab:image-manip}. We see that for both the scaling and shifting, both \MODELNearest and \MODELAngles approaches perform better than the vanilla LayoutLMv3 baseline for all cases. As expected, we see a drop in performance in all the models. We measure the difference in F-1 scores between results without manipulation (Table \ref{tab:lmv3-results}) and with manipulation. From these numbers, we see that for all models, the difference for each few-shot size is lower for both our approaches than the baselines for both FUNSD and CORD. This shows that our method is more robust compared to the baseline to these manipulations.

\subsection{Case studies}
We visualize several cases from the 4-shot setting. In Figure \ref{fig:lmv2-cases}, the models use LayoutLMv2 as backbone and we show an example from the FUNSD test set. We observe that compared to the ground-truth, the vanilla LayoutLMv2 makes errors specifically when there are a sequence of tokens next to each other all with the \texttt{I-ANSWER} tag. We see that both the approaches \MODELNearest and \MODELAngles are able to capture the continuous set of words in the form correctly. We believe that our graph based approach is able to capture the spatial relationship of these words and is thus able to get better predictions. Further, the \MODELAngles approach also captures the mislabeled \texttt{I-QUESTION} tags by \MODELNearest.
We show another example in Figure \ref{fig:lmv3-cases} in which we use LayoutLMv3 as the backbone and show an example from the CORD test set. We see that compared to the ground truth, the vanilla LayoutLMv3 model misclassifies the \texttt{TOTAL-CHANGEPRICE} tag. We see that both the approaches \MODELNearest and \MODELAngles are able to classify that correctly.



\section{Conclusion and Future Work}
We present \MODEL, a layout-aware graph based entity recognition framework for few-shot entity recognition in document images. Existing methods use the coordinates of the token bounding boxes to encode layout information and they are sensitive to manipulations in the images such as shifting, rotation or scaling especially in low data resource settings. Our approach makes use of the topological relationship between the tokens in the documents by using a graph-based approach and it is more robust to these manipulations. We construct graphs based on heuristics relating to the k-nearest neighbors of these tokens in space and at a certain angle. We extend layout-aware pre-trained language models with a graph attention network with the graphs we construct and the output hidden states froms the backbone language model. Extensive experiments in few-shot settings on FUNSD and CORD datasets illustrate the performance gains using our approach. Further, we show experiments with image manipulations where our approach is robust to these alterations in the image.
In the future, we plan to apply the model on other backbones and incorporate other features such as the semantic relationship between the tokens in addition to the topological relationship when constructing the graph.

\section*{Limitations}

The density of the graph constructed in terms of edge connectivity is dependent on the layout of the tokens present in the document. This leads to certain types of documents or even certain documents within a dataset to have a very dense graph whilst other documents can have sparse graphs. This could be a factor that affects the output representations from the GAT and the performance of the model.

\section*{Ethics Statement}
Our work focuses on few-shot entity recognition in document images. Both the datasets that we use are public and builds upon language models that are open-source. We also plan to release our code publicly. Thus, we do not anticipate any ethical concerns. 

 \clearpage


\clearpage
\nocite{*}
\section{Bibliographical References}\label{sec:reference}

\bibliographystyle{lrec-coling2024-natbib}
\bibliography{lrec-coling2024-example}

\begin{thebibliography}{43}
\expandafter\ifx\csname natexlab\endcsname\relax\def\natexlab#1{#1}\fi

\bibitem[{Aho and Ullman(1972)}]{Aho:72}
Alfred~V. Aho and Jeffrey~D. Ullman. 1972.
\newblock \emph{The Theory of Parsing, Translation and Compiling}, volume~1.
\newblock Prentice-Hall, Englewood Cliffs, NJ.

\bibitem[{{American Psychological Association}(1983)}]{APA:83}
{American Psychological Association}. 1983.
\newblock \emph{Publications Manual}.
\newblock American Psychological Association, Washington, DC.

\bibitem[{Ando and Zhang(2005)}]{Ando2005}
Rie~Kubota Ando and Tong Zhang. 2005.
\newblock \href {https://www.jmlr.org/papers/volume6/ando05a/ando05a.pdf} {A
  framework for learning predictive structures from multiple tasks and
  unlabeled data}.
\newblock \emph{Journal of Machine Learning Research}, 6:1817--1853.

\bibitem[{Andrew and Gao(2007)}]{andrew2007scalable}
Galen Andrew and Jianfeng Gao. 2007.
\newblock \href {https://dl.acm.org/doi/abs/10.1145/1273496.1273501} {Scalable
  training of {$L_1$}-regularized log-linear models}.
\newblock In \emph{Proceedings of the 24th International Conference on Machine
  Learning}, pages 33--40.

\bibitem[{Appalaraju et~al.(2021)Appalaraju, Jasani, Kota, Xie, and
  Manmatha}]{docformer2021}
Srikar Appalaraju, Bhavan Jasani, Bhargava~Urala Kota, Yusheng Xie, and
  R.~Manmatha. 2021.
\newblock \href {https://doi.org/10.48550/ARXIV.2106.11539} {Docformer:
  End-to-end transformer for document understanding}.

\bibitem[{BSI(1973{\natexlab{a}})}]{bs-2570-manual}
BSI. 1973{\natexlab{a}}.
\newblock \emph{Natural Fibre Twines}, 3rd edition.
\newblock British Standards Institution, London.
\newblock BS 2570.

\bibitem[{BSI(1973{\natexlab{b}})}]{bs-2570-techreport}
BSI. 1973{\natexlab{b}}.
\newblock Natural fibre twines.
\newblock BS 2570, British Standards Institution, London.
\newblock 3rd. edn.

\bibitem[{Castor and Pollux(1992)}]{CastorPollux-92}
A.~Castor and L.~E. Pollux. 1992.
\newblock The use of user modelling to guide inference and learning.
\newblock \emph{Applied Intelligence}, 2(1):37--53.

\bibitem[{Chandra et~al.(1981)Chandra, Kozen, and Stockmeyer}]{Chandra:81}
Ashok~K. Chandra, Dexter~C. Kozen, and Larry~J. Stockmeyer. 1981.
\newblock \href {https://doi.org/10.1145/322234.322243} {Alternation}.
\newblock \emph{Journal of the Association for Computing Machinery},
  28(1):114--133.

\bibitem[{Chercheur(1994)}]{Chercheur-94}
J.L. Chercheur. 1994.
\newblock \emph{Case-Based Reasoning}, 2nd edition.
\newblock Morgan Kaufman Publishers, San Mateo, CA.

\bibitem[{Chomsky(1973)}]{chomsky-73}
N.~Chomsky. 1973.
\newblock Conditions on transformations.
\newblock In \emph{A festschrift for {Morris Halle}}, New York. Holt, Rinehart
  \& Winston.

\bibitem[{Cooley and Tukey(1965)}]{ct1965}
James~W. Cooley and John~W. Tukey. 1965.
\newblock \href
  {https://www.ams.org/journals/mcom/1965-19-090/S0025-5718-1965-0178586-1/S0025-5718-1965-0178586-1.pdf}
  {An algorithm for the machine calculation of complex {F}ourier series}.
\newblock \emph{Mathematics of Computation}, 19(90):297--301.

\bibitem[{Devlin et~al.(2018)Devlin, Chang, Lee, and
  Toutanova}]{devlin2018bert}
Jacob Devlin, Ming-Wei Chang, Kenton Lee, and Kristina Toutanova. 2018.
\newblock Bert: Pre-training of deep bidirectional transformers for language
  understanding.
\newblock \emph{arXiv preprint arXiv:1810.04805}.

\bibitem[{Dong et~al.(2019)Dong, Yang, Wang, Wei, Liu, Wang, Gao, Zhou, and
  Hon}]{unilm}
Li~Dong, Nan Yang, Wenhui Wang, Furu Wei, Xiaodong Liu, Yu~Wang, Jianfeng Gao,
  Ming Zhou, and Hsiao-Wuen Hon. 2019.
\newblock Unified language model pre-training for natural language
  understanding and generation.
\newblock \emph{arXiv preprint arXiv:1905.03197}.

\bibitem[{Eco(1990)}]{Eco:1990}
Umberto Eco. 1990.
\newblock \emph{The Limits of Interpretation}.
\newblock Indian University Press.

\bibitem[{Guillaume~Jaume(2019)}]{jaume2019}
Jean-Philippe~Thiran Guillaume~Jaume, Hazim Kemal~Ekenel. 2019.
\newblock Funsd: A dataset for form understanding in noisy scanned documents.
\newblock In \emph{Accepted to ICDAR-OST}.

\bibitem[{Gusfield(1997)}]{Gusfield:97}
Dan Gusfield. 1997.
\newblock \href
  {https://www.cambridge.org/core/books/algorithms-on-strings-trees-and-sequences/F0B095049C7E6EF5356F0A26686C20D3}
  {\emph{Algorithms on Strings, Trees and Sequences}}.
\newblock Cambridge University Press, Cambridge, UK.

\bibitem[{Hoel(1971{\natexlab{a}})}]{hoel-71-whole}
Paul~Gerhard Hoel. 1971{\natexlab{a}}.
\newblock \emph{Elementary Statistics}, 3rd edition.
\newblock Wiley series in probability and mathematical statistics. Wiley, New
  York, Chichester.
\newblock ISBN 0~471~40300.

\bibitem[{Hoel(1971{\natexlab{b}})}]{hoel-71-portion}
Paul~Gerhard Hoel. 1971{\natexlab{b}}.
\newblock \emph{Elementary Statistics}, 3rd edition, Wiley series in
  probability and mathematical statistics, pages 19--33. Wiley, New York,
  Chichester.
\newblock ISBN 0~471~40300.

\bibitem[{Huang et~al.(2022)Huang, Lv, Cui, Lu, and Wei}]{huang2022layoutlmv3}
Yupan Huang, Tengchao Lv, Lei Cui, Yutong Lu, and Furu Wei. 2022.
\newblock Layoutlmv3: Pre-training for document ai with unified text and image
  masking.
\newblock In \emph{Proceedings of the 30th ACM International Conference on
  Multimedia}.

\bibitem[{Jespersen(1922)}]{Jespersen:1922}
Otto Jespersen. 1922.
\newblock \emph{Language: Its Nature, Development, and Origin}.
\newblock Allen and Unwin.

\bibitem[{Lee et~al.(2022)Lee, Li, Dozat, Perot, Su, Hua, Ainslie, Wang, Fujii,
  and Pfister}]{Lee2022FormNetSE}
Chen-Yu Lee, Chun-Liang Li, Timothy Dozat, Vincent Perot, Guolong Su, Nan Hua,
  Joshua Ainslie, Renshen Wang, Yasuhisa Fujii, and Tomas Pfister. 2022.
\newblock Formnet: Structural encoding beyond sequential modeling in form
  document information extraction.
\newblock In \emph{Annual Meeting of the Association for Computational
  Linguistics}.

\bibitem[{Lee et~al.(2023)Lee, Li, Zhang, Dozat, Perot, Su, Zhang, Sohn,
  Glushnev, Wang et~al.}]{lee2023formnetv2}
Chen-Yu Lee, Chun-Liang Li, Hao Zhang, Timothy Dozat, Vincent Perot, Guolong
  Su, Xiang Zhang, Kihyuk Sohn, Nikolai Glushnev, Renshen Wang, et~al. 2023.
\newblock Formnetv2: Multimodal graph contrastive learning for form document
  information extraction.
\newblock \emph{arXiv preprint arXiv:2305.02549}.

\bibitem[{Liu et~al.(2019)Liu, Ott, Goyal, Du, Joshi, Chen, Levy, Lewis,
  Zettlemoyer, and Stoyanov}]{liu2019roberta}
Yinhan Liu, Myle Ott, Naman Goyal, Jingfei Du, Mandar Joshi, Danqi Chen, Omer
  Levy, Mike Lewis, Luke Zettlemoyer, and Veselin Stoyanov. 2019.
\newblock Roberta: A robustly optimized bert pretraining approach.
\newblock \emph{arXiv preprint arXiv:1907.11692}.

\bibitem[{Lockard et~al.(2019)Lockard, Shiralkar, and
  Dong}]{Lockard2019OpenCeresWO}
Colin Lockard, Prashant Shiralkar, and Xin Dong. 2019.
\newblock Openceres: When open information extraction meets the semi-structured
  web.
\newblock In \emph{North American Chapter of the Association for Computational
  Linguistics}.

\bibitem[{Lockard et~al.(2020)Lockard, Shiralkar, Dong, and
  Hajishirzi}]{lockard2020zeroshotceres}
Colin Lockard, Prashant Shiralkar, Xin~Luna Dong, and Hannaneh Hajishirzi.
  2020.
\newblock \href {https://doi.org/10.18653/v1/2020.acl-main.721}
  {{Z}ero{S}hot{C}eres: Zero-shot relation extraction from semi-structured
  webpages}.
\newblock In \emph{Proceedings of the 58th Annual Meeting of the Association
  for Computational Linguistics}, pages 8105--8117, Online. Association for
  Computational Linguistics.

\bibitem[{Mandivarapu et~al.(2021)Mandivarapu, bunch, and
  fung}]{mandivarapu2021}
Jaya~Krishna Mandivarapu, Eric bunch, and Glenn fung. 2021.
\newblock \href {https://doi.org/10.48550/ARXIV.2111.00007} {Domain agnostic
  few-shot learning for document intelligence}.

\bibitem[{Marquez et~al.(2005)Marquez, Comas, Gim{\'e}nez, and
  Catala}]{marquez2005semantic}
Lluis Marquez, Pere Comas, Jes{\'u}s Gim{\'e}nez, and Neus Catala. 2005.
\newblock Semantic role labeling as sequential tagging.
\newblock In \emph{Proceedings of the Ninth Conference on Computational Natural
  Language Learning (CoNLL-2005)}, pages 193--196.

\bibitem[{Nakayama(2018)}]{seqeval}
Hiroki Nakayama. 2018.
\newblock \href {https://github.com/chakki-works/seqeval} {{seqeval}: A python
  framework for sequence labeling evaluation}.
\newblock Software available from https://github.com/chakki-works/seqeval.

\bibitem[{Park et~al.(2019)Park, Shin, Lee, Lee, Surh, Seo, and
  Lee}]{park2019cord}
Seunghyun Park, Seung Shin, Bado Lee, Junyeop Lee, Jaeheung Surh, Minjoon Seo,
  and Hwalsuk Lee. 2019.
\newblock Cord: A consolidated receipt dataset for post-ocr parsing.

\bibitem[{Rasooli and Tetreault(2015)}]{rasooli-tetrault-2015}
Mohammad~Sadegh Rasooli and Joel~R. Tetreault. 2015.
\newblock \href {http://arxiv.org/abs/1503.06733} {Yara parser: {A} fast and
  accurate dependency parser}.
\newblock \emph{Computing Research Repository}, arXiv:1503.06733.
\newblock Version 2.

\bibitem[{Ratinov and Roth(2009)}]{ratinov2009design}
Lev Ratinov and Dan Roth. 2009.
\newblock Design challenges and misconceptions in named entity recognition.
\newblock In \emph{Proceedings of the Thirteenth Conference on Computational
  Natural Language Learning (CoNLL-2009)}, pages 147--155.

\bibitem[{Singer et~al.(1954--58)Singer, Holmyard, and Hall}]{singer-whole}
Charles~Joseph Singer, E.~J. Holmyard, and A.~R. Hall, editors. 1954--58.
\newblock \emph{A history of technology}.
\newblock Oxford University Press, London.
\newblock 5 vol.

\bibitem[{Strötgen and Gertz(2012)}]{Martin-90}
Jannik Strötgen and Michael Gertz. 2012.
\newblock Temporal tagging on different domains: Challenges, strategies, and
  gold standards.
\newblock In \emph{Proceedings of the Eight International Conference on
  Language Resources and Evaluation (LREC'12)}, pages 3746--3753, Istanbul,
  Turkey. European Language Resource Association (ELRA).

\bibitem[{Superman et~al.(2000)Superman, Batman, Catwoman, and
  Spiderman}]{Superman-Batman-Catwoman-Spiderman-00}
S.~Superman, B.~Batman, C.~Catwoman, and S.~Spiderman. 2000.
\newblock \emph{Superheroes experiences with books}, 20th edition.
\newblock The Phantom Editors Associates, Gotham City.

\bibitem[{Vaswani et~al.(2017)Vaswani, Shazeer, Parmar, Uszkoreit, Jones,
  Gomez, Kaiser, and Polosukhin}]{vaswani2017attn}
Ashish Vaswani, Noam Shazeer, Niki Parmar, Jakob Uszkoreit, Llion Jones,
  Aidan~N Gomez, \L~ukasz Kaiser, and Illia Polosukhin. 2017.
\newblock \href
  {https://proceedings.neurips.cc/paper/2017/file/3f5ee243547dee91fbd053c1c4a845aa-Paper.pdf}
  {Attention is all you need}.
\newblock In \emph{Advances in Neural Information Processing Systems},
  volume~30. Curran Associates, Inc.

\bibitem[{Veli{\v{c}}kovi{\'{c}} et~al.(2018)Veli{\v{c}}kovi{\'{c}}, Cucurull,
  Casanova, Romero, Li{\`{o}}, and Bengio}]{velickovic2018graph}
Petar Veli{\v{c}}kovi{\'{c}}, Guillem Cucurull, Arantxa Casanova, Adriana
  Romero, Pietro Li{\`{o}}, and Yoshua Bengio. 2018.
\newblock \href {https://openreview.net/forum?id=rJXMpikCZ} {{Graph Attention
  Networks}}.
\newblock \emph{International Conference on Learning Representations}.

\bibitem[{Wang et~al.(2022)Wang, Gu, Tensmeyer, Barmpalios, Nenkova, Sun,
  Shang, and Morariu}]{wang2022mgdoc}
Zilong Wang, Jiuxiang Gu, Chris Tensmeyer, Nikolaos Barmpalios, Ani Nenkova,
  Tong Sun, Jingbo Shang, and Vlad~I. Morariu. 2022.
\newblock \href {https://doi.org/10.48550/ARXIV.2211.14958} {Mgdoc:
  Pre-training with multi-granular hierarchy for document image understanding}.

\bibitem[{Wang and Shang(2022)}]{wang2022laser}
Zilong Wang and Jingbo Shang. 2022.
\newblock \href {https://doi.org/10.18653/v1/2022.findings-acl.329} {Towards
  few-shot entity recognition in document images: A label-aware
  sequence-to-sequence framework}.
\newblock In \emph{Findings of the Association for Computational Linguistics:
  ACL 2022}, Dublin, Ireland. Association for Computational Linguistics.

\bibitem[{Wolf et~al.(2019)Wolf, Debut, Sanh, Chaumond, Delangue, Moi, Cistac,
  Rault, Louf, Funtowicz et~al.}]{wolf2019huggingface}
Thomas Wolf, Lysandre Debut, Victor Sanh, Julien Chaumond, Clement Delangue,
  Anthony Moi, Pierric Cistac, Tim Rault, R{\'e}mi Louf, Morgan Funtowicz,
  et~al. 2019.
\newblock Huggingface's transformers: State-of-the-art natural language
  processing.
\newblock \emph{arXiv preprint arXiv:1910.03771}.

\bibitem[{Xu et~al.(2021)Xu, Xu, Lv, Cui, Wei, Wang, Lu, Florencio, Zhang, Che,
  Zhang, and Zhou}]{Xu2020LayoutLMv2MP}
Yang Xu, Yiheng Xu, Tengchao Lv, Lei Cui, Furu Wei, Guoxin Wang, Yijuan Lu,
  Dinei Florencio, Cha Zhang, Wanxiang Che, Min Zhang, and Lidong Zhou. 2021.
\newblock Layoutlmv2: Multi-modal pre-training for visually-rich document
  understanding.
\newblock In \emph{Proceedings of the 59th Annual Meeting of the Association
  for Computational Linguistics (ACL) 2021}.

\bibitem[{Xu et~al.(2020)Xu, Li, Cui, Huang, Wei, and Zhou}]{xu2020layoutlm}
Yiheng Xu, Minghao Li, Lei Cui, Shaohan Huang, Furu Wei, and Ming Zhou. 2020.
\newblock Layoutlm: Pre-training of text and layout for document image
  understanding.
\newblock In \emph{Proceedings of the 26th ACM SIGKDD International Conference
  on Knowledge Discovery \& Data Mining}, pages 1192--1200.

\bibitem[{Zhang et~al.(2020)Zhang, Zhu, Hou, Liu, Yang, Wang, and
  Yin}]{Zhang_2020_CVPR}
Shi-Xue Zhang, Xiaobin Zhu, Jie-Bo Hou, Chang Liu, Chun Yang, Hongfa Wang, and
  Xu-Cheng Yin. 2020.
\newblock Deep relational reasoning graph network for arbitrary shape text
  detection.
\newblock In \emph{IEEE/CVF Conference on Computer Vision and Pattern
  Recognition (CVPR)}.

\end{thebibliography}


\section{Appendix}
\label{sec:appendix}

\subsection{Baseline models}
There are several popular open source models in this domain: re BERT \cite{devlin2018bert}, RoBERTa \cite{liu2019roberta}, LayoutLM\cite{xu2020layoutlm}, LayoutLMv2\cite{Xu2020LayoutLMv2MP} and LayoutLMv3\cite{huang2022layoutlmv3}. 
\begin{itemize}
    \item \textbf{BERT}\cite{devlin2018bert} is a text-only auto-encoding pre-trianed language model that uses masked language modeling and next sentence prediction as its pre-training tasks. For this task, we fine-tune the pre-trained BERT base model with the few-shot training samples for both the datasets.
    \item \textbf{RoBERTa}\cite{liu2019roberta} is an extension of BERT that is trained on more data and also makes modifications to its pre-training tasks thereby achieving better performance in numerous natural language understanding tasks. Similar to BERT, we fine-tune the base model with the few-shot training samples for both the datasets.
    \item \textbf{LayoutLM}\cite{xu2020layoutlm} is a multimodal language model that includes layout and text information. LayoutLM is built upon BERT and adds extra spatial embeddings into the BERT embedding layer. 
    \item \textbf{LayoutLMv2}\cite{Xu2020LayoutLMv2MP} is a multi-modal language model which is an improved version of LayoutLM~\citep{xu2020layoutlm}. It integrates the visual information in the pre-training stage to learn the cross-modality interaction between visual and textual information. 
    \item \textbf{LayoutLMv3}\cite{huang2022layoutlmv3} is another large multi-modal language model which aims to mitigate the discrepancy between text and image modalities in other models such as LayoutLM and LayoutLMv2. It facilitates multimodal representation learning by unifying the text and image masking.
\end{itemize}

From Table \ref{tab:all_baselines} we can see that LayoutLMv2 and LayoutLMv3 are the two strongest models and significantly outperform BERT, RoBERTa and LayoutLM. Thus, in our \MODEL framework, we perform our experiments by picking LayoutLmv2 or LayoutLMv3 as our pre-trained layout aware backbone language model.

Table \ref{tab:lmv2_all_results} and \ref{tab:lmv3_all_results} comprises of few-shot experimental results using LayoutLMv2 and LayoutLMv3 as the backbone models respectively for few-shot sizes from 1 to 10.

\begin{table}[h]
    \resizebox{\linewidth}{!}{
\begin{tabular}{@{}c|c|c@{}}
\toprule
\textbf{Model} & \textbf{FUNSD-F1} & \textbf{CORD-F1} \\ \midrule
LayoutLMv3 & 89.92 & 81.95 \\ \midrule
LayoutLMv3 + \MODELNearest & 90.23 & \textbf{82.14} \\ \midrule
LayoutLMv3 + \MODELAngles & \textbf{90.34} & 82.09
\end{tabular}}

\caption{Evaluation results with LayoutLMv3 as the backbone model using the entire dataset. \textbf{Bold} denotes the best model}
\label{tab:full_dataset}
\end{table}

\begin{table*}[t]
\centering

\resizebox{\linewidth}{!}{
\setlength{\tabcolsep}{4mm}{
\begin{tabular}{cccccccc}
\toprule
\multirow{2}{*}{\textbf{$\mathcal{P}$}} & \multirow{2}{*}{\textbf{Model}} & \multicolumn{3}{c}{\textbf{FUNSD}} & \multicolumn{3}{c}{\textbf{CORD}} \\
 &  & \multicolumn{1}{c}{\textbf{Precision}} & \multicolumn{1}{c}{\textbf{Recall}} & \multicolumn{1}{c}{\textbf{F-1}} & \multicolumn{1}{c}{\textbf{Precision}} & \multicolumn{1}{c}{\textbf{Recall}} & \multicolumn{1}{c}{\textbf{F-1}} \\ \hline
\multirow{5}{*}{1} & BERT & 10.93$\pm$3.63 & 22.39$\pm$6.68 & 13.97$\pm$2.84 & 22.4$\pm$4.91 & 31.24$\pm$6.4 & 26.08$\pm$5.55 \\
 & RoBERTa & 12.98$\pm$4.34 & 21.93$\pm$11.09 & 17.13$\pm$6.07 & 18.55$\pm$6.69 & 25.77$\pm$9.24 & 21.56$\pm$7.73 \\
 & LayoutLM & 15.77$\pm$5.35 & 23.03$\pm$7.52 & 22.03$\pm$5.15 & 26.5$\pm$9.06 & 35.58$\pm$11.17 & 30.37$\pm$10.02 \\
 & LayoutLMv2 & \textbf{23.4$\pm$10.05} & \textbf{29.22$\pm$14.6} & \textbf{25.51$\pm$11.72} & \textbf{32.5$\pm$4.15} & \textbf{42.51$\pm$3.82} & \textbf{36.33$\pm$4.03} \\
 & LayoutLMv3 & \textbf{22.93$\pm$6.21} & \textbf{39.7$\pm$5.94} & \textbf{28.85$\pm$6.35} & \textbf{36.56$\pm$5.46} & \textbf{47.58$\pm$6.44} & \textbf{41.33$\pm$5.88} \\ \hline
\multirow{5}{*}{2} & BERT & 15.51$\pm$2.29 & 28.14$\pm$4.02 & 19.74$\pm$2.05 & 30.05$\pm$5.93 & 41.63$\pm$6.02 & 34.87$\pm$6.12 \\
 & RoBERTa & 21.64$\pm$1.64 & 33.43$\pm$4.24 & 26.68$\pm$1.76 & 34.96$\pm$6.73 & 45.7$\pm$7.17 & 39.59$\pm$7.03 \\
 & LayoutLM & 33.05$\pm$4.85 & 35.52$\pm$8.81 & 28.02$\pm$6.07 & 38.51$\pm$7.88 & 50.52$\pm$6.81 & 43.63$\pm$7.66 \\
 & LayoutLMv2 & \textbf{38.61$\pm$4.04} & \textbf{53.8$\pm$8.35} & \textbf{44.82$\pm$5.3} & \textbf{43.34$\pm$2.07} & \textbf{55.85$\pm$2.43} & \textbf{48.66$\pm$2.13} \\
 & LayoutLMv3 & \textbf{44.29$\pm$6.14} & \textbf{58.96$\pm$7.2} & \textbf{50.43$\pm$6.03} & \textbf{47.21$\pm$6.25} & \textbf{58.99$\pm$4.94} & \textbf{52.41$\pm$5.85} \\ \hline
\multirow{5}{*}{3} & BERT & 19.42$\pm$3.75 & 32.63$\pm$5.62 & 24.3$\pm$4.44 & 32.57$\pm$8.07 & 44.9$\pm$8.73 & 37.72$\pm$8.53 \\
 & RoBERTa & 25.22$\pm$3.22 & 39.0$\pm$5.37 & 30.57$\pm$3.76 & 41.0$\pm$8.37 & 51.07$\pm$8.35 & 45.46$\pm$8.48 \\
 & LayoutLM & 28.69$\pm$3.86 & 46.07$\pm$8.95 & 35.13$\pm$7.29 & 43.35$\pm$6.77 & 56.15$\pm$4.62 & 48.84$\pm$6.11 \\
 & LayoutLMv2 & \textbf{47.73$\pm$2.91} & \textbf{61.16$\pm$7.02} & \textbf{53.55$\pm$4.36} & \textbf{50.52$\pm$2.87} & \textbf{61.03$\pm$3.8} & \textbf{54.69$\pm$3.19} \\
 & LayoutLMv3 & \textbf{59.66$\pm$4.92} & \textbf{72.2$\pm$7.65} & \textbf{65.29$\pm$5.92} & \textbf{51.34$\pm$6.55} & \textbf{62.49$\pm$5.47} & \textbf{56.34$\pm$6.2} \\ \hline
\multirow{5}{*}{4} & BERT & 21.2$\pm$3.54 & 37.04$\pm$3.13 & 26.9$\pm$3.59 & 36.48$\pm$8.43 & 48.17$\pm$8.47 & 41.47$\pm$8.64 \\
 & RoBERTa & 27.53$\pm$2.92 & 42.83$\pm$2.68 & 33.48$\pm$2.83 & 45.89$\pm$7.84 & 55.04$\pm$8.69 & 50.05$\pm$8.25 \\
 & LayoutLM & 34.31$\pm$2.56 & 52.23$\pm$5.45 & 41.29$\pm$2.68 & 48.41$\pm$6.28 & 60.5$\pm$4.25 & 53.7$\pm$5.58 \\
 & LayoutLMv2 & \textbf{51.13$\pm$2.16} & \textbf{64.57$\pm$6.8} & \textbf{56.95$\pm$3.7} & \textbf{55.18$\pm$2.25} & \textbf{65.96$\pm$3.35} & \textbf{59.88$\pm$2.66} \\
 & LayoutLMv3 & \textbf{65.32$\pm$3.89} & \textbf{77.97$\pm$2.26} & \textbf{71.06$\pm$3.04} & \textbf{54.18$\pm$5.01} & \textbf{64.92$\pm$3.76} & \textbf{59.04$\pm$4.53} \\ \hline
\multirow{5}{*}{5} & BERT & 24.2$\pm$3.24 & 39.59$\pm$2.55 & 29.97$\pm$3.0 & 37.75$\pm$8.26 & 49.53$\pm$8.19 & 42.81$\pm$8.37 \\
 & RoBERTa & 31.57$\pm$2.56 & 46.77$\pm$2.14 & 37.65$\pm$2.2 & 48.51$\pm$8.28 & 57.32$\pm$10.11 & 52.54$\pm$9.07 \\
 & LayoutLM & 38.60$\pm$5.12 & 54.07$\pm$5.49 & 44.87$\pm$4.61 & 52.05$\pm$5.68 & 63.7$\pm$3.79 & 57.23$\pm$4.95 \\
 & LayoutLMv2 & \textbf{53.46$\pm$2.43} & \textbf{64.57$\pm$6.71} & \textbf{58.36$\pm$3.63} & \textbf{57.59$\pm$3.11} & \textbf{68.11$\pm$3.24} & \textbf{62.27$\pm$3.06} \\
 & LayoutLMv3 & \textbf{67.14$\pm$5.17} & \textbf{77.88$\pm$2.62} & \textbf{72.07$\pm$4.01} & \textbf{58.55$\pm$2.82} & \textbf{67.03$\pm$2.46} & \textbf{62.49$\pm$2.57} \\ \hline
\multirow{5}{*}{6} & BERT & 26.54$\pm$1.99 & 41.47$\pm$3.69 & 32.27$\pm$1.99 & 42.04$\pm$5.46 & 54.18$\pm$4.43 & 47.31$\pm$5.13 \\
 & RoBERTa & 33.75$\pm$2.19 & 47.2$\pm$2.54 & 39.32$\pm$2.06 & 52.88$\pm$4.84 & 61.41$\pm$4.86 & 56.82$\pm$4.82 \\
 & LayoutLM & 42.27$\pm$3.85 & 57.84$\pm$4.49 & 48.79$\pm$3.75 & 52.05$\pm$5.68 & 63.7$\pm$3.79 & 57.23$\pm$4.95 \\
 & LayoutLMv2 & \textbf{56.92$\pm$1.59} & \textbf{67.59$\pm$2.9} & \textbf{61.77$\pm$1.77} & \textbf{60.28$\pm$2.47} & \textbf{70.01$\pm$2.48} & \textbf{64.41$\pm$2.43} \\
 & LayoutLMv3 & \textbf{71.19$\pm$3.75} & \textbf{80.83$\pm$1.09} & \textbf{75.68$\pm$2.58} & \textbf{60.91$\pm$3.51} & \textbf{69.16$\pm$2.76} & \textbf{64.76$\pm$3.16} \\ \hline
\multirow{5}{*}{7} & BERT & 28.67$\pm$1.94 & 44.04$\pm$4.07 & 34.59$\pm$1.64 & 43.31$\pm$5.46 & 54.8$\pm$4.5 & 48.35$\pm$5.15 \\
 & RoBERTa & 35.27$\pm$2.7 & 49.24$\pm$4.49 & 41.01$\pm$2.77 & 54.95$\pm$3.64 & 62.66$\pm$4.0 & 58.55$\pm$3.78 \\
 & LayoutLM & 45.81$\pm$2.59 & 61.24$\pm$4.05 & 52.29$\pm$1.93 & 57.9$\pm$2.22 & 67.88$\pm$1.96 & 62.48$\pm$1.95 \\
 & LayoutLMv2 & \textbf{59.43$\pm$3.59} & \textbf{68.98$\pm$3.73} & \textbf{63.71$\pm$2.19} & \textbf{60.71$\pm$2.09} & \textbf{69.95$\pm$2.16} & \textbf{64.86$\pm$2.03} \\
 & LayoutLMv3 & \textbf{72.44$\pm$3.56} & \textbf{81.56$\pm$1.12} & \textbf{76.68$\pm$1.95} & \textbf{62.26$\pm$3.78} & \textbf{70.3$\pm$3.02} & \textbf{66.03$\pm$3.43} \\ \hline

\multirow{5}{*}{8} & BERT & 30.81$\pm$2.83 & 43.72$\pm$3.99 & 36.11$\pm$3.12 & 45.58$\pm$5.27 & 57.34$\pm$4.27 & 50.76$\pm$4.94 \\
 & RoBERTa & 37.3$\pm$3.55 & 49.52$\pm$4.89 & 42.52$\pm$3.93 & 57.38$\pm$1.86 & 65.32$\pm$1.54 & 61.08$\pm$1.57 \\
 & LayoutLM & 48.48$\pm$3.09 & 60.21$\pm$4.66 & 53.68$\pm$3.51 & 57.9$\pm$2.22 & 67.88$\pm$1.96 & 62.48$\pm$1.95 \\
 & LayoutLMv2 & \textbf{61.69$\pm$2.93} & \textbf{69.9$\pm$3.3} & \textbf{65.51$\pm$2.75} & \textbf{62.98$\pm$0.94} & \textbf{72.07$\pm$1.24} & \textbf{67.18$\pm$1.07} \\
 & LayoutLMv3 & \textbf{74.31$\pm$2.19} & \textbf{81.75$\pm$2.6} & \textbf{77.85$\pm$2.29} & \textbf{64.49$\pm$3.24} & \textbf{72.21$\pm$2.17} & \textbf{68.12$\pm$2.77} \\ \hline

\multirow{5}{*}{9} & BERT & 31.18$\pm$2.75 & 43.67$\pm$5.27 & 36.33$\pm$3.51 & 47.25$\pm$3.93 & 59.11$\pm$2.9 & 52.5$\pm$3.54 \\
 & RoBERTa & 37.3$\pm$3.41 & 49.74$\pm$4.26 & 42.6$\pm$3.61 & 58.77$\pm$2.22 & 66.52$\pm$1.09 & 62.39$\pm$1.62 \\
 & LayoutLM & 51.91$\pm$2.37 & 63.59$\pm$4.04 & 57.14$\pm$2.95 & 57.9$\pm$2.22 & 67.88$\pm$1.96 & 62.48$\pm$1.95 \\
 & LayoutLMv2 & \textbf{62.54$\pm$2.22} & \textbf{71.17$\pm$3.65} & \textbf{66.55$\pm$2.54} & \textbf{63.93$\pm$0.5} & \textbf{72.63$\pm$0.63} & \textbf{67.96$\pm$0.15} \\
 & LayoutLMv3 & \textbf{75.9$\pm$1.53} & \textbf{82.52$\pm$1.36} & \textbf{79.06$\pm$1.24} & \textbf{65.89$\pm$2.8} & \textbf{73.44$\pm$1.87} & \textbf{69.45$\pm$2.39} \\ \hline

\multirow{5}{*}{10} & BERT & 32.32$\pm$3.55 & 45.16$\pm$5.04 & 37.59$\pm$3.84 & 50.83$\pm$3.11 & 61.91$\pm$2.25 & 55.81$\pm$2.76 \\
 & RoBERTa & 38.65$\pm$3.64 & 51.1$\pm$4.77 & 43.98$\pm$3.93 & 60.22$\pm$2.28 & 67.73$\pm$1.59 & 63.75$\pm$1.9 \\
 & LayoutLM & 52.94$\pm$2.51 & 64.29$\pm$3.3 & 58.05$\pm$2.63 & 64.05$\pm$2.76 & 71.99$\pm$1.91 & 67.78$\pm$2.33 \\
 & LayoutLMv2 & \textbf{63.49$\pm$2.7} & \textbf{72.97$\pm$2.21} & \textbf{67.89$\pm$2.33} & \textbf{66.18$\pm$0.99} & \textbf{73.79$\pm$0.8} & \textbf{69.92$\pm$0.76} \\
 & LayoutLMv3 & \textbf{75.9$\pm$1.53} & \textbf{82.52$\pm$1.36} & \textbf{79.06$\pm$1.24} & \textbf{65.89$\pm$2.8} & \textbf{73.44$\pm$1.87} & \textbf{69.45$\pm$2.39} \\
 \bottomrule
\end{tabular}
}
}
\caption{Comparison of various baseline models for all few-shot sizes. \textbf{Bold} denotes the best two models}
\label{tab:all_baselines}
\end{table*}

\begin{table*}[t]
\centering

\resizebox{\linewidth}{!}{
    \setlength{\tabcolsep}{4mm}{
\begin{tabular}{lccllllll}
\hline
\multirow{2}{*}{} & \multicolumn{1}{l}{\multirow{2}{*}{\textbf{$\mathcal{P}$}}} & \multirow{2}{*}{\textbf{Model}} & \multicolumn{3}{c}{\textbf{FUNSD}} & \multicolumn{3}{c}{\textbf{CORD}} \\
 & \multicolumn{1}{l}{} &  & \multicolumn{1}{c}{\textbf{Precision}} & \multicolumn{1}{c}{\textbf{Recall}} & \multicolumn{1}{c}{\textbf{F-1}} & \multicolumn{1}{c}{\textbf{Precision}} & \multicolumn{1}{c}{\textbf{Recall}} & \multicolumn{1}{c}{\textbf{F-1}} \\ \hline
\multirow{30}{*}{\rotatebox[origin=c]{90}{LayoutLMv2}} & \multirow{3}{*}{1} & Vanilla & 23.4$\pm$10.05 & 29.22$\pm$14.6 & 25.51$\pm$11.72 & 32.5$\pm$4.15 & 42.51$\pm$3.82 & 36.33$\pm$4.03 \\
 &  & + \MODELNearest & \textbf{26.66$\pm$4.36} & \textbf{40.56$\pm$7.7} & \textbf{32.08$\pm$5.35} & \textbf{31.75$\pm$6.3} & \textbf{43.55$\pm$7.56} & \textbf{36.36$\pm$6.89} \\
 &  & + \MODELAngles & 26.58$\pm$5.06 & 36.62$\pm$13.16 & 29.72$\pm$8.13 & 30.56$\pm$6.63 & 41.37$\pm$8.34 & 35.14$\pm$7.39 \\ \cline{2-9} 
 & \multirow{3}{*}{2} & Vanilla & 38.61$\pm$4.04 & 53.8$\pm$8.35 & 44.82$\pm$5.3 & 43.34$\pm$2.07 & 55.85$\pm$2.43 & 48.66$\pm$2.13 \\
 &  & + \MODELNearest & 41.4$\pm$3.26 & 52.08$\pm$9.89 & 45.74$\pm$5.32 & 43.42$\pm$2.45 & 55.89$\pm$3.15 & 48.87$\pm$2.73 \\
 &  & + \MODELAngles & \textbf{41.45$\pm$2.22} & \textbf{54.21$\pm$2.99} & \textbf{46.9$\pm$1.66} & \textbf{43.77$\pm$2.77} & \textbf{56.22$\pm$2.94} & \textbf{49.21$\pm$2.8} \\ \cline{2-9} 
 & \multirow{3}{*}{3} & Vanilla & 47.73$\pm$2.91 & 61.16$\pm$7.02 & 53.55$\pm$4.36 & 50.52$\pm$2.87 & 61.03$\pm$3.8 & 54.69$\pm$3.19 \\
 &  & + \MODELNearest & \textbf{48.72$\pm$3.33} & \textbf{63.12$\pm$8.69} & \textbf{54.9$\pm$5.42} & \textbf{50.81$\pm$3.03} & \textbf{61.49$\pm$3.48} & \textbf{55.02$\pm$3.09} \\
 &  & + \MODELAngles & 48.48$\pm$3.46 & 60.68$\pm$9.14 & 53.79$\pm$5.73 & 50.15$\pm$2.79 & 61.28$\pm$2.98 & 54.53$\pm$2.7 \\ \cline{2-9} 
 & \multirow{3}{*}{4} & Vanilla & 51.13$\pm$2.16 & 64.57$\pm$6.8 & 56.95$\pm$3.7 & 55.18$\pm$2.25 & 65.96$\pm$3.35 & 59.88$\pm$2.66 \\
 &  & + \MODELNearest & \textbf{52.59$\pm$3.71} & \textbf{67.23$\pm$4.75} & \textbf{58.93$\pm$3.55} & 54.86$\pm$2.57 & 65.87$\pm$3.88 & 59.85$\pm$3.05 \\
 &  & + \MODELAngles & 50.86$\pm$3.43 & 66.47$\pm$4.56 & 57.59$\pm$3.63 & \textbf{55.52$\pm$2.56} & \textbf{66.57$\pm$3.45} & \textbf{60.54$\pm$2.89} \\ \cline{2-9}
 & \multirow{3}{*}{5} & Vanilla & 53.46$\pm$2.43 & 64.57$\pm$6.71 & 58.36$\pm$3.63 & 57.59$\pm$3.11 & 68.11$\pm$3.24 & 62.27$\pm$3.06 \\
 &  & + \MODELNearest & \textbf{54.0$\pm$2.54} & \textbf{67.77$\pm$4.98} & \textbf{60.02$\pm$2.83} & 57.49$\pm$3.1 & 68.27$\pm$2.55 & 62.41$\pm$2.82 \\
 &  & + \MODELAngles & 52.98$\pm$3.25 & 67.23$\pm$5.08 & 59.19$\pm$3.53 & \textbf{57.88$\pm$2.73} & \textbf{68.27$\pm$2.48} & \textbf{62.63$\pm$2.52} \\ \cline{2-9}
 & \multirow{3}{*}{6} & Vanilla & 56.92$\pm$1.59 & 67.59$\pm$2.9 & 61.77$\pm$1.77 & 60.28$\pm$2.47 & 70.01$\pm$2.48 & 64.41$\pm$2.43 \\
 &  & + \MODELNearest & 57.45$\pm$2.57 & 71.14$\pm$2.83 & 63.53$\pm$2.25 & \textbf{60.63$\pm$2.88} & \textbf{70.49$\pm$2.69} & \textbf{65.19$\pm$2.78} \\
 &  & + \MODELAngles & \textbf{58.14$\pm$3.79} & \textbf{71.34$\pm$3.02} & \textbf{63.95$\pm$2.21} & 59.97$\pm$2.37 & 70.04$\pm$2.28 & 64.6$\pm$2.26 \\ \cline{2-9} 
 & \multirow{3}{*}{7} & Vanilla & 59.43$\pm$3.59 & 68.98$\pm$3.73 & 63.71$\pm$2.19 & 60.71$\pm$2.09 & 69.95$\pm$2.16 & 64.86$\pm$2.03 \\
 &  & + \MODELNearest & 59.13$\pm$3.96 & 70.83$\pm$3.18 & 64.33$\pm$2.45 & \textbf{60.87$\pm$2.04} & 70.38$\pm$1.86 & \textbf{65.27$\pm$1.79} \\
 &  & + \MODELAngles & \textbf{61.11$\pm$3.53} & \textbf{72.16$\pm$2.74} & \textbf{66.13$\pm$2.76} & 60.84$\pm$1.5 & 70.31$\pm$1.76 & 65.23$\pm$1.46 \\ \cline{2-9} 
 & \multirow{3}{*}{8} & Vanilla & 61.69$\pm$2.93 & 69.9$\pm$3.3 & 65.51$\pm$2.75 & 62.98$\pm$0.94 & 72.07$\pm$1.24 & 67.18$\pm$1.07 \\
 &  & + \MODELNearest & \textbf{63.31$\pm$2.0} & 71.99$\pm$1.87 & \textbf{67.35$\pm$1.46} & \textbf{63.07$\pm$2.31} & 72.08$\pm$2.47 & 67.27$\pm$2.34 \\
 &  & + \MODELAngles & 61.92$\pm$4.21 & \textbf{72.4$\pm$2.85} & 66.63$\pm$2.5 & 63.28$\pm$1.75 & \textbf{72.23$\pm$2.0} & \textbf{67.46$\pm$1.84} \\ \cline{2-9}
 & \multirow{3}{*}{9} & Vanilla & 62.54$\pm$2.22 & 71.17$\pm$3.65 & 66.55$\pm$2.54 & 63.93$\pm$0.5 & 72.63$\pm$0.63 & 67.96$\pm$0.15 \\
 &  & + \MODELNearest & 62.62$\pm$2.47 & 71.26$\pm$2.1 & 66.64$\pm$1.95 & \textbf{64.19$\pm$2.53} & \textbf{72.99$\pm$1.21} & \textbf{68.3$\pm$1.88} \\
 &  & + \MODELAngles & \textbf{63.53$\pm$3.81} & \textbf{71.94$\pm$2.64} & \textbf{67.44$\pm$3.08} & 63.72$\pm$1.75 & 72.79$\pm$1.29 & 67.94$\pm$1.4 \\ \cline{2-9} 
 & \multirow{3}{*}{10} & Vanilla & 63.49$\pm$2.7 & 72.17$\pm$2.21 & 67.59$\pm$2.33 & 66.18$\pm$0.99 & 73.79$\pm$0.8 & 69.92$\pm$0.76 \\
 &  & + \MODELNearest & \textbf{63.9$\pm$2.87} & 71.89$\pm$3.03 & \textbf{67.63$\pm$2.62} & 66.09$\pm$1.34 & 74.0$\pm$0.5 & 69.82$\pm$0.9 \\
 &  & + \MODELAngles & 62.77$\pm$3.98 & \textbf{72.71$\pm$1.67} & 67.33$\pm$2.78 & \textbf{66.37$\pm$1.24} & \textbf{74.19$\pm$0.69} & \textbf{70.06$\pm$0.95} \\ \hline
\end{tabular}
}
}
\caption{Evaluation results with LayoutLMv2 as baseline on all few-shot sizes. \textbf{Bold} indicates best model}
\label{tab:lmv2_all_results}
\end{table*}

\begin{table*}[t]

\centering
\resizebox{\linewidth}{!}{
    \setlength{\tabcolsep}{4mm}{
\begin{tabular}{lccllllll}
\toprule
\multirow{2}{*}{} & \multicolumn{1}{l}{\multirow{2}{*}{\textbf{$\mathcal{P}$}}} & \multirow{2}{*}{\textbf{Model}} & \multicolumn{3}{c}{\textbf{FUNSD}} & \multicolumn{3}{c}{\textbf{CORD}} \\
 & \multicolumn{1}{l}{} &  & \multicolumn{1}{c}{\textbf{Precision}} & \multicolumn{1}{c}{\textbf{Recall}} & \multicolumn{1}{c}{\textbf{F-1}} & \multicolumn{1}{c}{\textbf{Precision}} & \multicolumn{1}{c}{\textbf{Recall}} & \multicolumn{1}{c}{\textbf{F-1}} \\ \hline
\multirow{30}{*}{\rotatebox[origin=c]{90}{LayoutLMv3}} & \multirow{3}{*}{1} & Vanilla & 22.93$\pm$6.21 & 39.7$\pm$5.94 & 28.85$\pm$6.35 & 36.56$\pm$5.46 & 47.58$\pm$6.44 & 41.33$\pm$5.88 \\
 &  & +  \MODELNearest & \textbf{28.74$\pm$8.4} & \textbf{35.99$\pm$7.71} & \textbf{31.49$\pm$7.63} & \textbf{38.87$\pm$6.57} & \textbf{49.83$\pm$6.76} & \textbf{43.64$\pm$6.73} \\
 &  & +  \MODELAngles & 26.11$\pm$6.69 & 32.53$\pm$6.29 & 28.23$\pm$5.24 & 37.52$\pm$6.34 & 48.7$\pm$6.6 & 42.36$\pm$6.54 \\ \cline{2-9} 
 & \multirow{3}{*}{2} & Vanilla & 44.29$\pm$6.14 & 58.96$\pm$7.2 & 50.43$\pm$6.03 & 47.21$\pm$6.25 & 58.99$\pm$4.94 & 52.41$\pm$5.85 \\
 &  & +   \MODELNearest & \textbf{49.82$\pm$6.06} & \textbf{59.55$\pm$8.91} & \textbf{54.09$\pm$6.54} & \textbf{48.68$\pm$5.72} & 60.19$\pm$4.23 & \textbf{53.79$\pm$5.24} \\
 &  & +  \MODELAngles & 46.8$\pm$6.46 & 58.15$\pm$8.92 & 51.61$\pm$6.42 & 48.15$\pm$5.07 & \textbf{60.3$\pm$3.57} & 53.51$\pm$4.58 \\ \cline{2-9} 
 & \multirow{3}{*}{3} & Vanilla & 59.66$\pm$4.92 & 72.2$\pm$7.65 & 65.29$\pm$5.92 & 51.34$\pm$6.55 & 62.49$\pm$5.47 & 56.34$\pm$6.2 \\
 &  & +  \MODELNearest & \textbf{62.18$\pm$5.13} & \textbf{73.12$\pm$7.3} & \textbf{67.12$\pm$5.56} & \textbf{53.08$\pm$7.32} & \textbf{64.3$\pm$5.55} & \textbf{58.1$\pm$6.73} \\
 &  & +  \MODELAngles & 60.73$\pm$5.09 & 72.41$\pm$7.64 & 65.97$\pm$5.71 & 52.77$\pm$7.17 & 63.72$\pm$5.55 & 57.68$\pm$6.63 \\ \cline{2-9} 
 & \multirow{3}{*}{4} & Vanilla & 65.32$\pm$3.89 & 77.97$\pm$2.26 & 71.06$\pm$3.04 & 54.18$\pm$5.01 & 64.92$\pm$3.76 & 59.04$\pm$4.53 \\
 &  & +   \MODELNearest & \textbf{67.86$\pm$3.3} & \textbf{78.73$\pm$2.57} & \textbf{72.86$\pm$2.69} & \textbf{56.28$\pm$4.24} & \textbf{66.47$\pm$3.29} & \textbf{60.94$\pm$3.86} \\
 &  & +  \MODELAngles & 65.93$\pm$3.28 & 77.22$\pm$3.45 & 71.08$\pm$2.81 & 55.38$\pm$4.63 & 65.99$\pm$3.79 & 60.21$\pm$4.3 \\ \cline{2-9} 
 & \multirow{3}{*}{5} & Vanilla & 67.14$\pm$5.17 & 77.88$\pm$2.62 & 72.07$\pm$4.01 & 58.55$\pm$2.82 & 67.03$\pm$2.46 & 62.49$\pm$2.57 \\
 &  & +  \MODELNearest & 69.6$\pm$2.57 & 79.72$\pm$1.66 & 74.3$\pm$1.94 & \textbf{59.84$\pm$3.27} & 68.36$\pm$2.34 & \textbf{63.8$\pm$2.76} \\
 &  & +  \MODELAngles & \textbf{70.32$\pm$1.41} & \textbf{80.86$\pm$1.23} & \textbf{75.22$\pm$1.1} & 59.37$\pm$4.09 & \textbf{68.48$\pm$3.08} & 63.58$\pm$3.63 \\ \cline{2-9} 
 & \multirow{3}{*}{6} & Vanilla & 71.19$\pm$3.75 & 80.83$\pm$1.09 & 75.68$\pm$2.58 & 60.91$\pm$3.51 & 69.16$\pm$2.76 & 64.76$\pm$3.16 \\
 &  & +  \MODELNearest & \textbf{72.71$\pm$3.42} & 81.53$\pm$1.98 & \textbf{76.84$\pm$2.58} & \textbf{61.8$\pm$5.14} & 70.0$\pm$3.75 & 65.63$\pm$4.53 \\
 &  & +  \MODELAngles & 72.31$\pm$3.7 & \textbf{81.65$\pm$1.81} & 76.67$\pm$2.7 & 61.56$\pm$4.49 & \textbf{70.3$\pm$3.33} & \textbf{65.63$\pm$3.98} \\ \cline{2-9} 
 & \multirow{3}{*}{7} & Vanilla & 72.44$\pm$3.56 & 81.56$\pm$1.12 & 76.68$\pm$1.95 & 62.26$\pm$3.78 & 70.3$\pm$3.02 & 66.03$\pm$3.43 \\
 &  & +  \MODELNearest & 74.48$\pm$2.42 & 82.4$\pm$1.1 & 78.22$\pm$1.49 & \textbf{62.84$\pm$4.17} & \textbf{71.05$\pm$3.26} & \textbf{66.68$\pm$3.77} \\
 &  & +  \MODELAngles & \textbf{74.63$\pm$2.69} & \textbf{83.18$\pm$2.04} & \textbf{78.65$\pm$2.06} & 62.77$\pm$4.17 & 71.01$\pm$3.15 & 66.62$\pm$3.72 \\ \cline{2-9} 
 & \multirow{3}{*}{8} & Vanilla & 74.31$\pm$2.19 & 81.75$\pm$2.6 & 77.85$\pm$2.29 & 64.49$\pm$3.24 & 72.21$\pm$2.17 & 68.12$\pm$2.77 \\
 &  & +  \MODELNearest & 76.27$\pm$1.44 & 83.41$\pm$1.73 & 79.66$\pm$1.14 & 64.89$\pm$4.38 & 72.22$\pm$3.19 & 68.35$\pm$3.84 \\
 &  & +  \MODELAngles & \textbf{76.41$\pm$2.18} & \textbf{83.98$\pm$1.49} & \textbf{79.99$\pm$1.24} & \textbf{65.1$\pm$4.14} & \textbf{72.27$\pm$2.87} & \textbf{68.49$\pm$3.56} \\ \cline{2-9} 
 & \multirow{3}{*}{9} & Vanilla & 75.9$\pm$1.53 & 82.52$\pm$1.36 & 79.06$\pm$1.24 & 65.89$\pm$2.8 & 73.44$\pm$1.87 & 69.45$\pm$2.39 \\
 &  & +  \MODELNearest & 76.83$\pm$1.95 & 83.23$\pm$1.61 & 79.89$\pm$1.51 & \textbf{66.84$\pm$3.25} & \textbf{73.62$\pm$1.85} & \textbf{70.05$\pm$2.6} \\
 &  & +  \MODELAngles & \textbf{76.95$\pm$2.03} & \textbf{84.43$\pm$1.65} & \textbf{80.5$\pm$1.59} & 66.73$\pm$3.99 & 73.45$\pm$2.35 & 69.91$\pm$3.25 \\ \cline{2-9} 
 & \multirow{3}{*}{10} & Vanilla & 76.1$\pm$2.31 & 82.65$\pm$2.34 & 79.24$\pm$2.27 & 66.72$\pm$2.65 & 73.58$\pm$2.08 & 69.97$\pm$2.34 \\
 &  & +  \MODELNearest & \textbf{77.44$\pm$2.65} & \textbf{84.43$\pm$1.83} & 80.77$\pm$2.02 & \textbf{67.68$\pm$3.49} & 73.64$\pm$2.57 & \textbf{70.53$\pm$3.05} \\
 &  & +  \MODELAngles & 77.39$\pm$2.53 & 84.57$\pm$1.53 & \textbf{80.81$\pm$1.79} & 67.05$\pm$3.2 & \textbf{73.64$\pm$1.93} & 70.18$\pm$2.61 \\ \bottomrule
\end{tabular}
}}
\caption{Evaluation results with LayoutLMv3 as baseline on all few-shot sizes. \textbf{Bold} indicates best model}
\label{tab:lmv3_all_results}
\end{table*}

\subsection{Experiments with entire datasets}
 Though the model that we construct isn’t tailored for a few-shot setting, we believe our approach is particularly useful in a few-shot setting when there is limited data availability for entity recognition, as the graph-based method is efficient to train and easier to generalize. We perform additional experiments using the entire dataset to validate if the graph based approach is effective even when trained with the entire dataset and not in a few-shot setting. We run experiments using LayoutLMv3 as the backbone as shown in Table \ref{tab:full_dataset} and observe improvements in both of our approaches, \MODELNearest and \MODELAngles.  However, our main contribution is for the few-shot setting. Our approach focuses on few-shot settings and limited data availability scenarios where we have potentially a larger number of documents for testing.



\end{document}